%% file: main.tex
\documentclass[10pt,twocolumn,letterpaper]{article}
\usepackage[pagenumbers]{style/cvpr} % To force page numbers, e.g. for an arXiv version

\usepackage{graphicx}
\usepackage{amsmath}
\usepackage{amssymb}
\usepackage{booktabs}
\usepackage[table]{xcolor}
\usepackage{comment}
\usepackage{letltxmacro}

\usepackage[pagebackref,breaklinks,colorlinks]{hyperref}

\usepackage[capitalize]{cleveref}
\crefname{section}{Sec.}{Secs.}
\Crefname{section}{Section}{Sections}
\Crefname{table}{Table}{Tables}
\crefname{table}{Tab.}{Tabs.}

\newcommand{\xhdr}[1]{\vspace{2pt}\noindent\textbf{#1}}

\def\cvprPaperID{****} % *** Enter the CVPR Paper ID here
\def\confName{CVPR}
\def\confYear{2022}

\LetLtxMacro{\oldsection}{\section}
\renewcommand{\section}[1]{
    \oldsection{#1}
    \vspace{-0.10in}
}

\LetLtxMacro{\oldsubsection}{\subsection}
\renewcommand{\subsection}[1]{
    \oldsubsection{#1}
    \vspace{-0.08in}
}

\begin{document}
\input{cover}
\input{sections/0_abstract}
\input{sections/1_introduction}
\input{sections/2_related}
\input{sections/3_dataset}
\input{sections/4_models}
\input{sections/5_baselines}
\input{sections/6_results}
\input{sections/7_conclusion}

%%%%%%%%% REFERENCES
\LetLtxMacro{\section}{\oldsection}
{\small
\bibliographystyle{style/ieee_fullname}

}

\pagebreak
\input{sections/appendix}

\end{document}

%% file: cover.tex
\title{Episodic Memory Question Answering}

\author{
Samyak Datta\\
Georgia Tech\\
\and
Sameer Dharur\\
Georgia Tech\\
\and
Vincent Cartillier\\
Georgia Tech\\
\and
Ruta Desai\\
Meta Reality Labs Research\\
\and
Mukul Khanna\\
\and
Dhruv Batra\\
Georgia Tech, Meta AI Research\\
\and
Devi Parikh\\
Georgia Tech, Meta AI Research
}
\maketitle

%% file: sections/0_abstract.tex
\begin{abstract}
   Egocentric augmented reality devices such as wearable glasses passively capture visual data as a human wearer tours a home environment. We envision a scenario wherein the human communicates with an AI agent powering such a device by asking questions (e.g., ``where did you last see my keys?''). In order to succeed at this task, the egocentric AI assistant must (1) construct semantically rich and efficient scene memories that encode spatio-temporal information about objects seen during the tour and (2) possess the ability to understand the question and ground its answer into the semantic memory representation. Towards that end, we introduce (1) a new task — Episodic Memory Question Answering (EMQA) wherein an egocentric AI assistant is provided with a video sequence (the tour) and a question as an input and is asked to localize its answer to the question within the tour, (2) a dataset of grounded questions designed to probe the agent’s spatio-temporal understanding of the tour, and (3) a model for the task that encodes the scene as an allocentric, top-down semantic feature map and grounds the question into the map to localize the answer. We show that our choice of episodic scene memory outperforms naive, off-the-shelf solutions for the task as well as a host of very competitive baselines and is robust to noise in depth, pose as well as camera jitter. The project page can be found at \href{https://samyak-268.github.io/emqa/}{https://samyak-268.github.io/emqa/}.
\end{abstract}

%% file: sections/1_introduction.tex
\section{Introduction}
\label{sec:intro}

Imagine wearing a pair of AI-powered, augmented reality (AR) glasses and walking around your house. Such smart-glasses will possess the ability to ``see'' and passively capture egocentric visual data from the same perspective as its wearer, organize the surrounding visual information into its memory, and use this encoded information to communicate with humans by answering questions such as, ``where did you last see my keys?''. In other words, such devices can act as our own personal egocentric AI assistants.

\begin{figure}
    \centering
    \includegraphics[width=0.475\textwidth]{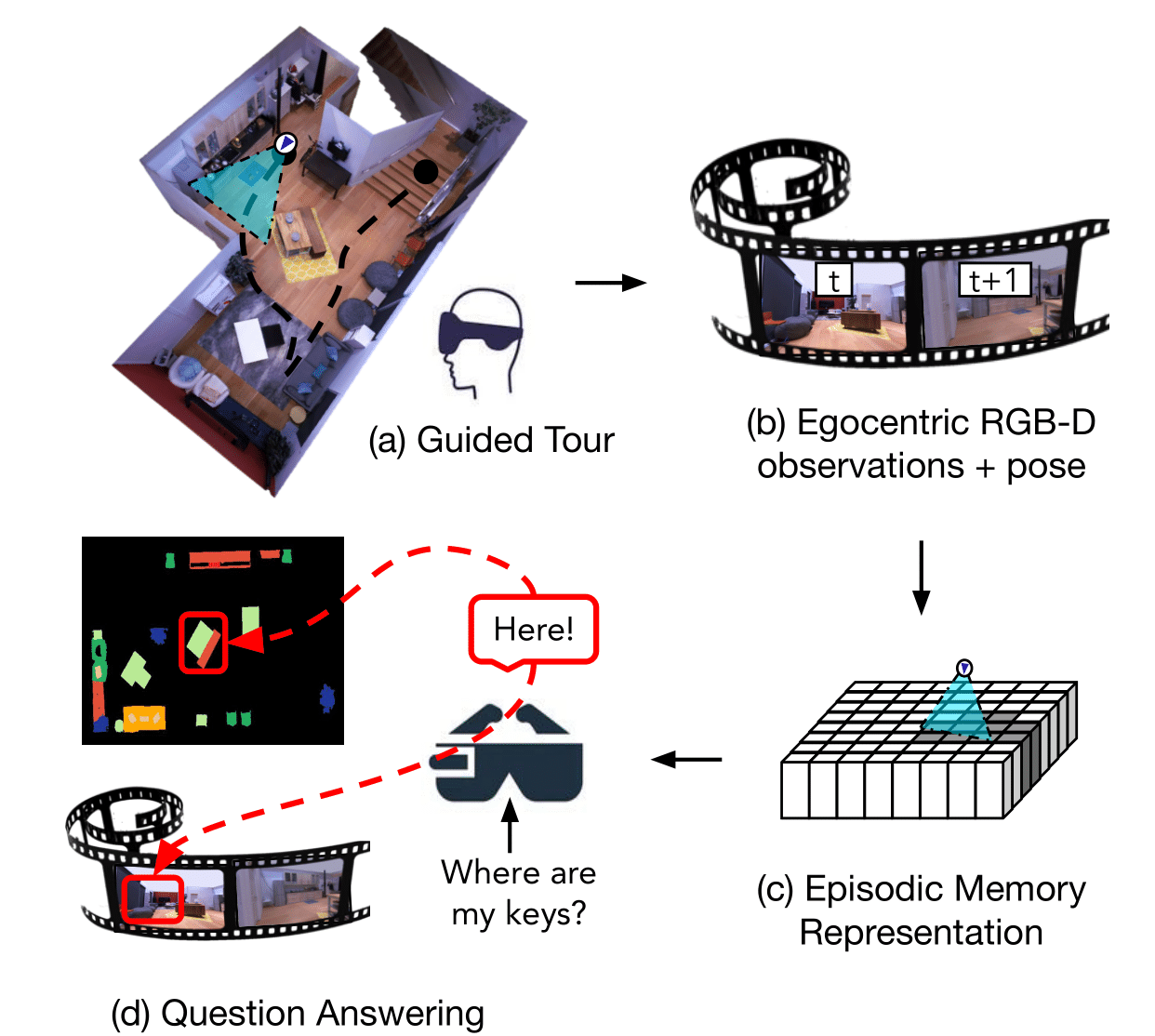}
    \caption{
        (a) An egocentric AI assistant, assumed to be running on a pair of augmented reality glasses, is taken on a guided exploratory tour by virtue of its human wearer moving about inside an environment (example scene from \cite{straub2019replica}). (b) The agent passively records an egocentric stream of RGB-D maps, (c) builds an internal, episodic memory representation of the scene and (d) exploits this spaito-temporal memory representation to answer (multiple) post-hoc questions about the tour.
    }
    \label{fig:teaser}
\end{figure}

There has been a rich history of prior work in training navigational agents to answer questions grounded in indoor environments -- a task referred to as Embodied Question Answering (EQA) in literature \cite{das2017embodied,yu2019multi,gordon2018iqa,wijmans2019embodied}. However, egocentric AI assistants differ from EQA agents in several important ways. First, such systems passively observe a sequence of egocentric visual frames as a result of the human wearer’s navigation, as opposed to taking actions in an environment. Second, AI systems for egocentric assistants would be required to build scene-specific memory representations that persist across different questions. This is in direct contrast with EQA where contemporary approaches have treated every question as a clean-slate navigation episode. EQA agents start navigating with no prior information about the scene (even if the current question is about a scene that they have witnessed before). And third, EQA agents respond to questions by uttering language token(s). Responding to a question such as, ``where did you last see my keys?'' with the answer — ``hallway'' isn’t a very helpful response if there are multiple hallways in the house. In contrast, our setup presents a scenario wherein an egocentric assistant can potentially localize answers by grounding them within the environment tour.

Therefore, as a step towards realizing the goal of such egocentric AI assistants, we present a novel task wherein the AI assistant is taken on a guided tour of an indoor environment and then asked to localize its answers to post-hoc questions grounded in the environment tour (Fig. \ref{fig:teaser}). This pre-exploratory tour presents an opportunity to build an internal, episodic memory of the scene. Once constructed, the AI assistant can utilize this scene memory to answer multiple, follow-up questions about the tour. We call this task -- Episodic Memory Question Answering (EMQA).

\begin{comment}
\begin{figure*}
    \centering
    \includegraphics[clip, trim=0cm 1.5cm 0cm 4cm, width=1.00\textwidth]{figures/teaser_with_headings.pdf}
    \caption{
        (a) An egocentric AI assistant, assumed to be running on a pair of augmented reality glasses, is taken on a guided exploratory tour by virtue of its human wearer moving about inside the environment. (b) The agent passively records an egocentric stream of RGB-D maps along with associated GT pose. (c) It builds an internal, episodic memory representation of the scene (what objects were seen, where were they observed and when). (d) The AI assistant later exploits this memory representation to answer (multiple) questions about the scene.
    }
    \label{fig:teaser}
\end{figure*}
\end{comment}

More concretely, in the proposed EMQA task, the system receives a pre-recorded sequence of RGB-D images with the corresponding oracle pose information (the guided agent tour) as an input. It uses the input tour to construct a memory representation of the indoor scene. Then, it exploits the scene memory to ground answers to multiple text questions. The localization of answers can happen either within the tour’s egocentric frame sequence or onto a top-down metric map (such as the house floorplan). The two output modalities are equivalent, given the agent pose.

The paper makes several important contributions. \textbf{First}, we introduce the task of Episodic Memory Question Answering. We generate a dataset of questions grounded in pre-recorded agent tours that are designed to probe the system’s spatial understanding of the scene ("where did you see the cushion?") as well as temporal reasoning abilities (“where did you first/last see the cushion?”). \textbf{Second}, we propose a model for the EMQA task that builds allocentric top-down semantic scene representations (the episodic scene memory) during the tour and leverages the same for answering follow-up questions. In order to build the episodic scene memory, our model combines the semantic features extracted from egocentric observations from the tour in a geometrically consistent manner into a single top-down feature map of the scene as witnessed during the tour \cite{cartillier2020}. \textbf{Third}, we extend existing scene memories that model spatial relationships among objects \cite{cartillier2020} (“what” objects were observed and “where") by augmenting them with temporal information (“when” were these objects observed), thereby making the memory amenable for reasoning about temporal localization questions.

\textbf{Fourth}, we compare our choice of scene representation against a host of baselines and show that our proposed model outperforms language-only baselines by $\sim150\%$, naive, “off-the-shelf” solutions to the task that rely on making frame-by-frame localization predictions by 37\% as well as memory representations from prior work that aggregate (via averaging \cite{eslami2018neural}, GRU \cite{cangea2019videonavqa}, context-conditioned attention \cite{fang2019scene}) buffers of observation features across the tour. 

\textbf{Finally}, in addition to photorealistic indoor environments \cite{Matterport3D}, we also test the robustness of our approach under settings that have a high fidelity to the real world. We show qualitative results of a zero-shot transfer of our approach to a real-world, RGB-D dataset \cite{sturm2012benchmark} that presents significantly challenging conditions of imperfect depth, pose and camera jitter - typical deployment conditions for egocentric AR assistants. In addition to that, we break away from the unrealistic assumption of the availability of oracle pose in indoor settings \cite{datta2020integrating, zhao2021surprising} and perform a systematic study of the impact of noise (of varying types and intensities) in the agent's pose. We show that our model is more resilient than baselines to such noisy perturbations.

%% file: sections/2_related.tex
\section{Related Work}
\label{sec:related}

\xhdr{Question-Answering in Embodied Environments}. 
Training embodied agents to answer questions grounded in indoor environments has been the subject of several prior works. Specifically, \cite{das2017embodied, yu2019multi} present agents that can answer questions about a single and multiple target objects in simulated scenes, respectively. \cite{wijmans2019embodied} presents an instantiation of this line of work to photorealistic scenes from the Matterport3D simulator and 3D point cloud based inputs whereas \cite{gordon2018iqa} presents an extension of the task that requires the agent to interact with its environment. In all of the above, the agents are setup to carry out each question-answering episode with a "clean-slate” i.e. with no avenues for the agent to potentially re-use scene information gathered during prior traversals of the same scene. Although the agent in \cite{gordon2018iqa} has a semantic spatial memory storing information about object semantics and free space, it is not persistent across different question-episodes from the same scene. Moreover, all prior work involves generating a ranked list of language answer tokens as predictions. Our task formulation allows for the sharing of a semantic scene memory that is persistent across the different questions from a scene and localizes answers to questions -- a much more high-fidelity setting for egocentric AI assistants.

\xhdr{Video Question Answering.}
Our work is also reminiscent of video question-answering (VideoQA) tasks. VideoQA has witnessed a rich history of prior work with the introduction of datasets sampled from "open-world” domains (movies \cite{tapaswi2016movieqa}, TV shows \cite{lei2018tvqa}, cooking recipes \cite{DaXuDoCVPR2013,zhou2017procnets}) and tasks involving detecting/localizing actions and answering questions about events that transpire in the videos. Our EMQA dataset instead, comprises of egocentric videos generated from navigation trajectories in indoor environments. In addition to localization within the input video sequence, this setting enables additional output modalities, such as grounding over scene floor-plans, that are incompatible with existing VideoQA domains. Moreover, existing VideoQA datasets are accompanied with rich, per-frame annotations such as subtitles, plot scripts and other sub-event metadata. In contrast, EMQA assumes no such additional annotations. \cite{grauman2021ego4d} is concurrent with our work and proposes a large-scale dataset of egocentric videos paired with 3D meshes of environments in which the recorded activities took place.

\xhdr{Localization via Dialog}. \cite{hahn2020you, de2018talk} present a task wherein an entity having access to the top-down floorplan of an environment needs to localize an agent (on the top-down map) navigating within a scene through first-person vision alone. Both assume that the top-down floorplan of the scene is provided to the agent as input whereas our model constructs the map representation from scratch from the guided tours.

\xhdr{Scene Memory Representations.}
Building representations of scenes for assisting agents in embodied tasks has a rich history of prior work with early examples adopting the hidden state of an LSTM navigation model as a compact representation of the scene \cite{wijmans2020ddppo, habitat19iccv}. To overcome the limited expressive power of a single state vector representing complex, 3D scenes, more recent approaches have modelled scene memories as either a buffer of observed egocentric features \cite{fang2019scene,eslami2018neural}, 2D metric grids \cite{cartillier2020,blukis2018mapping,Anderson2019ChasingGI,chaplot2020object}, topological maps \cite{savinov2018semi,wu2019bayesian} or full-scale 3D semantic maps \cite{tung2019learning,cheng2018geometry,prabhudesai2019embodied}. Storing observed egocentric features in a buffer does not explicitly model the geometric and spatial relationships among objects in the scene. Topological graphs are not optimal for precise metric localization (a desideratum for our task). The memory constraints involved with constructing voxel-based 3D scene representations restricts the feature maps proposed in \cite{tung2019learning,cheng2018geometry} to simple scenes (handful of objects placed on a table-top). On the contrary, our task deals with indoor environments of significantly higher visual complexity. Our scene memory (allocentric, top-down semantic feature map) is most similar to \cite{cartillier2020}. We present a novel extension to the scene features from \cite{cartillier2020} by incorporating temporal information with the semantic features (``when'' was ``what'' observed and ``where'').

%% file: sections/3_dataset.tex
\section{EMQA Dataset: Questions Grounded in Scene Tours}
\label{sec:dataset-questions}

We now describe the dataset for the task. Recall that the task involves taking the assistant on an exploratory tour of an indoor environment, and then asking multiple, post-hoc questions about the guided tour. The EMQA model must localize answers to the questions in the scene. 

\xhdr{Guided Exploration Tours}
\label{subsec:tours}
We instantiate our task in the Habitat \cite{habitat19iccv} simulator using Matterport3D \cite{Matterport3D} (MP3D) scans (reconstructed 3D meshes from 90 indoor environments). For any given indoor scene, we use the manually recorded exploration paths from \cite{cartillier2020}. These multi-room navigation trajectories were optimized for coverage, comprise of egocentric RGB-D maps, ground-truth pose and are, on an average, 2500 steps in length.

\xhdr{Questions Grounded in Tours}
\label{subsec:dataset-questions}
For a given exploration path through an indoor scene, we now describe our process of generating grounded questions about objects witnessed by the egocentric assistant. Following \cite{cartillier2020}, we restrict ourselves to $12$ commonly occurring object categories such as sofa, bed etc. (see Suppl. for the full list). 

We start by generating the ground-truth top-down maps labelled with object instances for each scene via an orthographic projection of the semantically annotated MP3D mesh. These generated maps contain ground-truth information about the top-down layout of all objects in all parts of a scene. Each cell in these maps is of a fixed spatial resolution of $2$cm $\times 2$cm and therefore, the spatial dimensions of the maps depend on the size of the indoor scenes. Although the exploration tours have been optimized for coverage, they do not cover all observable parts of all scenes (leaving out some hard-to-reach, niche areas of environments during the manually guided exploration process). Therefore, to ensure the relevance of questions in our dataset, next, we compute the subset of ``observed'' locations from within the ground-truth  semantic map of the entire scene.

\begin{figure*}
    \centering
    \includegraphics[clip, trim=0cm 0cm 0cm 0cm, scale=0.20]{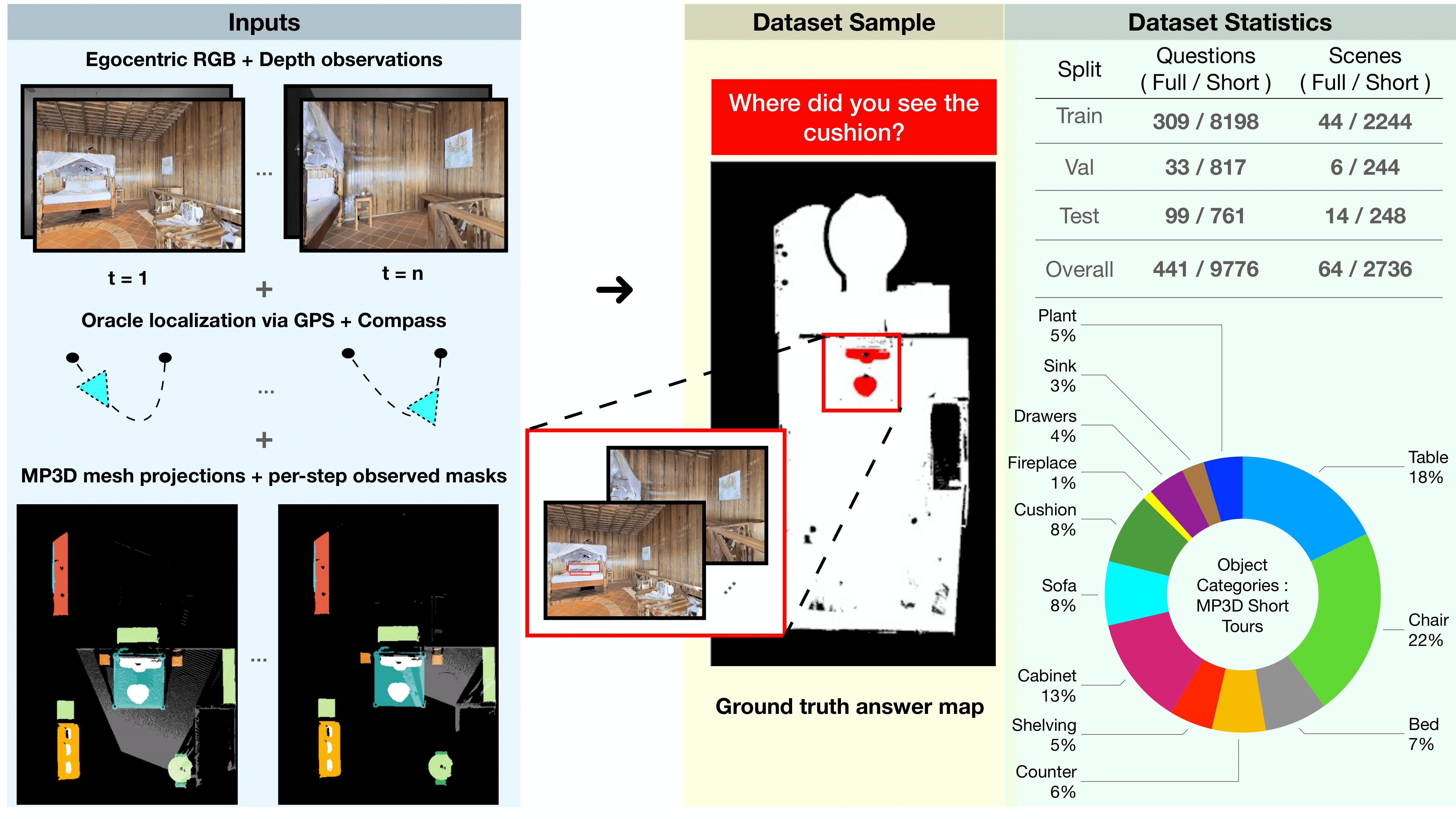}
    \caption{A schematic overview of our dataset generation process. The ground-truth top-down maps are created from the orthographic projection of the Matterport \cite{Matterport3D} mesh with semantic labels, via egocentric RGB + Depth observations and pose information, and filtered out by the per-step observed masks.}
    \label{fig:dataset}
\end{figure*}

In order to do that, we project the depth maps from each step of the tour onto the top-down scene map, giving us the per-time step mask over locally observed locations (Fig. \ref{fig:dataset}, ``Inputs''). Summing across all time steps and overlaying the resulting mask over the ground-truth semantic map, gives us the subset of objects and their locations observed during the tour. This serves as our source for generating questions.

From each such ``observed'' top-down semantic map (representing the tour), we generate templated questions that broadly belong to the following two categories: (1) \emph{spatial} localization questions and (2) \emph{spatio-temporal} localization questions. For the former, we generate a question of the form, ``where did you see the $<$X$>$?", where $<$X$>$ is an object category (from the pre-selected vocabulary of 12 objects) with at most 5 instances witnessed during the tour. Along with every question, we also record the information regarding (a) top-down map pixels corresponding to all instances of the object category in question that serves as the ground-truth answer, (b) the tour time-steps when each answer instance was observed during the tour. This is done by computing the intersection between the per-step mask of observed locations (described above) with the object instance on the top-down map. If, at any time step, the observed mask covers more than a heuristically determined fraction (10\%) of the object instance in question, then, we consider that instance to be ``seen'' at that particular time step of the tour (Fig. \ref{fig:dataset}, ``Dataset Sample''). 

Similarly, for the spatio-temporal subset, we generate questions with the format, ``where did you first/last see the $<$X$>$?" for each object category $<$X$>$ with atleast 2 instances sighted during the tour. To select the first (or, the last) viewed instance of the object, we query the the metadata (described above) comprising of the time steps wherein each instance of the object was viewed during the tour. The instance with the earliest (latest) time step among the first (last) observed time steps of all instances becomes the first (last) viewed instance of the object. Note that, in certain situations, the first and last seen instance of an object might coincide (the tour comprising of a ``loop'' within its tour), presenting a challenging scenario for learning.

\xhdr{Generating ``short" tours for training}
As stated above, the exploration tours in the EMQA dataset are, on an average, 2500 steps in length. To make the memory requirements and speed during training tractable, we follow the protocol laid down in \cite{cartillier2020} and consider 20-step ``short'' tour segments randomly sampled from the originally curated ``full'' tours. The top-down maps encompassing the area covered by these 20-step tour subsets are of fixed spatial dimensions of $250 \times 250$ cells across all ``short'' tours (as compared to varying spatial dimensions, depending on the environment size for ``full'' tours). We use the same question generation engine (described in the preceding sub-section) to generate questions corresponding to the ``short'' tour data splits. In addition to easing up speed and memory requirements during training, this also greatly increases the number of training samples to learn from. We would like to emphasize that our original task (and all results that follow) are defined on the full-scale tours.

Fig. \ref{fig:dataset} (``Dataset Statistics'') shows the distribution of the total number of scenes and questions across train, val and test splits for both ``short’’ and ``full’’ tours. We use mutually exclusive scenes for the train, val and test splits, evaluating generalization on never-before-seen environments. We also show a distribution of object categories across all questions in our dataset. For more qualitative examples %of the dataset 
and statistics (analysis of the sizes and spatial distribution of objects in scenes), please refer to the Suppl. document.

%% file: sections/4_models.tex
\section{Models}
\label{sec:models}

\begin{figure*}
    \centering
    \includegraphics[clip, trim=0cm 0cm 0cm 0cm, scale=0.50]{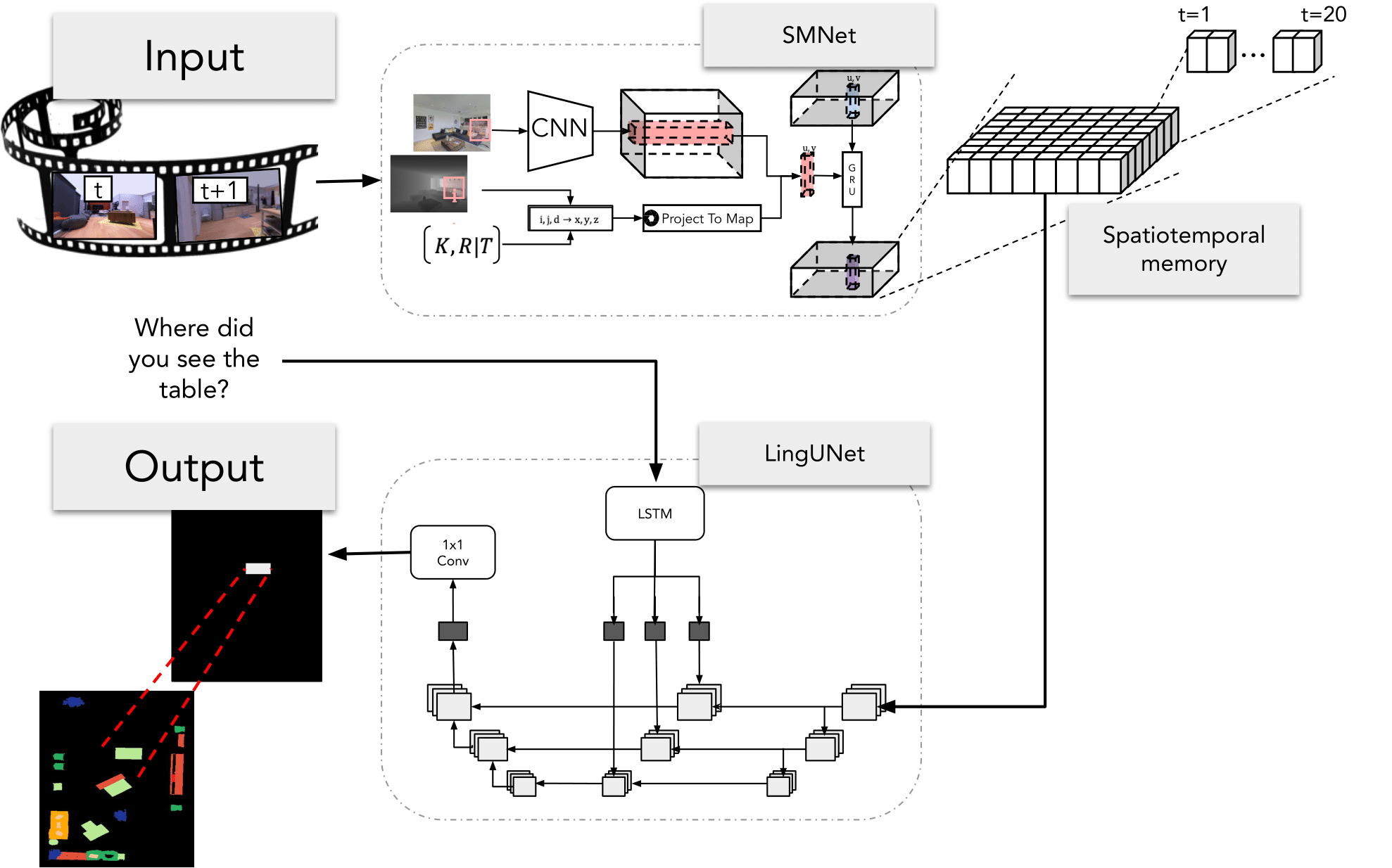}
    \vspace{0.1in}
    \caption{A schematic diagram of our proposed EMQA agent. Our agent first constructs an episodic memory representation of the tour and then grounds answers to questions on the top-down scene floorplan using a LingUNet-based answering model. A sample tour from the Matterport \cite{Matterport3D} is shown in the figure.}
    \label{fig:model}
\end{figure*}

Any model for the EMQA task must, broadly speaking, comprise of modules for the following two sub-tasks: (1) scene memory representation and (2) question-answering. The former takes the scene tour (RGB-D video frame sequence and associated ground-truth pose) as an input and generates a compact representation of the scene that ideally encodes information about objects, their relative spatial arrangements and when they were viewed during the tour. The latter takes the question as an input, operates on this episodic scene memory and generates its predicted answer as the output. In this section, we describe our choices regarding the specific instantiations of the two modules.

\xhdr{Scene Memory Representation.}
\label{subsec:scene-memory}
As our preferred choice of scene representation, we use allocentric, 2D, top-down, semantic features \cite{cartillier2020}. These representations are computed via projecting egocentric visual features onto an allocentric, top-down floor-plan of the environment using the knowledge of camera pose and depth.

\begin{table*}[h]
    \rowcolors{2}{gray!10}{white}
    \small
    \centering
        \begin{tabular}{l c c  c c c c c c}
        \toprule
        && \multicolumn{3}{c}{\textbf{{\footnotesize Top-down map output space}}} && \multicolumn{3}{c}{\textbf{{\footnotesize Egocentric pixel output space}}}  \\
        \cmidrule{3-5} \cmidrule{7-9}
        \textbf{{\footnotesize Method}} && {\footnotesize IoU} & {\footnotesize Recall} & {\footnotesize Precision} && {\footnotesize IoU} & {\footnotesize Recall} &  {\footnotesize Precision} \\
        \cmidrule{1-9}
        LangOnly && 4.75 {\tiny $\pm$ 0.14} & 14.41 {\tiny $\pm$ 0.62} & 6.98 {\tiny $\pm$ 0.18} && 5.26 {\tiny $\pm$ 0.18} & 14.57 {\tiny $\pm$ 0.62} & 7.86 {\tiny $\pm$ 0.19} \\
        EgoSemSeg && 22.89 {\tiny $\pm$ 0.69} & 35.85 {\tiny $\pm$ 1.06} & 38.45 {\tiny $\pm$ 1.07} &&23.42 {\tiny $\pm$ 0.73} & 36.06 {\tiny $\pm$ 1.06} & 40.13 {\tiny $\pm$ 1.15} \\
        SMNetDecoder && 26.92 {\tiny $\pm$ 1.12} & 43.86 {\tiny $\pm$ 1.25} & \textbf{40.95} {\tiny $\pm$ 1.26} &&27.13 {\tiny $\pm$ 1.12} & 43.39 {\tiny $\pm$ 1.58} & \textbf{41.96} {\tiny $\pm$ 1.22} \\

        \hline
        EgoBuffer-Avg \cite{eslami2018neural} && 0.07 {\tiny $\pm$ 0.01} & 0.37 {\tiny $\pm$ 0.06} & 0.24 {\tiny $\pm$ 0.03} && 0.11 {\tiny $\pm$ 0.01} & 0.40 {\tiny $\pm$ 0.06} & 0.34 {\tiny $\pm$ 0.03} \\
        EgoBuffer-GRU \cite{bakker2001reinforcement} && 0.01 {\tiny $\pm$ 0.00} & 0.01 {\tiny $\pm$ 0.00} & 0.02 {\tiny $\pm$ 0.00}  && 0.01 {\tiny $\pm$ 0.00} & 0.01 {\tiny $\pm$ 0.00} & 0.04 {\tiny $\pm$ 0.00}\\
        EgoBuffer-Attn \cite{fang2019scene} && 0.12 {\tiny $\pm$ 0.02} & 0.16 {\tiny $\pm$ 0.02} & 0.85 {\tiny $\pm$ 0.15} && 0.14 {\tiny $\pm$ 0.03} & 0.17 {\tiny $\pm$ 0.03} & 1.01 {\tiny $\pm$ 0.17} \\
        \hline
        
        \textbf{Ours} && 27.42 {\tiny $\pm$ 0.64} & 60.81 {\tiny $\pm$ 1.28} & 31.94 {\tiny $\pm$ 0.60} && 28.04 {\tiny $\pm$ 0.94} & 60.96 {\tiny $\pm$ 1.41} & 32.83 {\tiny $\pm$ 0.96}\\
        \textbf{Ours} (+temporal) && \textbf{29.11} {\tiny $\pm$ 0.44} & \textbf{62.27} {\tiny $\pm$ 1.13} & 33.39 {\tiny $\pm$ 0.51} && \textbf{29.78} {\tiny $\pm$ 0.59} & \textbf{62.68} {\tiny $\pm$ 1.08} & 34.36 {\tiny $\pm$ 0.73} \\
        \bottomrule
        \end{tabular}
    \\ [8pt]
    \caption{EMQA results for our proposed model and baselines in the ``top-down map'' and ``egocentric pixel'' output space.}
    \label{tab:results-topdown}
\end{table*}

More specifically, as shown in Fig. \ref{fig:model} (SMNet), at each step of the tour, we first extract convolutional features from the input RGB-D video frame via a RedNet \cite{jiang2018rednet} model that has been trained for egocentric semantic segmentation on indoor scenes from the SUN-RGBD \cite{sunrgb} dataset and then fine-tuned on egocentric frames from Matterport3D.
Next, these per-step egocentric semantic features are projected onto the scene's top-down floor-plan. The resulting feature map is of a fixed resolution -- each ``cell'' in the map corresponds to a fixed, $2$cm $\times 2$cm metric space in the real world and encodes semantic information about the objects present in that space (as viewed top-down). These local, per-step, projected features from each time step are accumulated using a GRU into a consolidated spatial memory tensor that serves as an ``episodic memory representation'' of the tour. The GRU is pre-trained to decode the top-down semantic segmentation from the scene memory tensor \cite{cartillier2020} with the ground truth top-down semantic map for the tour being computed from the annotated Matterport3D semantic mesh, as described in Sec. \ref{sec:dataset-questions}.

 On account of how these 2D scene features are derived, they have greater expressive capacity, model inter-object spatial relationships better (due to geometrically consistent pin-hole feature projections) and do not suffer from memory constraints of voxel-based representations.

\xhdr{Spatiotemporal memory.}
While the above is a sensible choice for representing objects in scenes, it doesn't encode temporal information about the tour (when were the objects observed?). Therefore, in this work, we present a novel extension to the scene representations from \cite{cartillier2020} by augmenting information about when each metric cell in the representation was observed during the tour.  
As shown in Fig. \ref{fig:model} (Spatiotemporal memory), this is done by a channel-wise stacking of the per-step masks over ``observed'' locations (from Sec. \ref{sec:dataset-questions}). Refer to the Supp. for more details.

\xhdr{Question-Answering.}
\label{subsec:question-answering}
We adopt the LingUNet \cite{blukis2018mapping} architecture to ground answers to input questions by exploiting the constructed scene memory (Fig. \ref{fig:model} (LingUNet)). LingUNet is an encoder-decoder architecture with language-conditioned skip connections. Prior work \cite{blukis2018mapping, hahn2020you, Anderson2019ChasingGI} has demonstrated that it is a highly performant architecture for tasks such as grounding of language-specified goal locations onto top-down maps of scenes for agent navigation.

The input questions are encoded using a 64-dim, single-layer LSTM. Our 3-layer LingUNet-based question-answering model takes the constructed scene memory features and the LSTM embedding of the question as inputs and generates a spatial feature map over the 2D floorplan (refer to Suppl. for the layer-wise architectural details of the LingUNet model). The spatial feature map is further processed by a convolutional block to generate the spatial distribution of answer predictions scores. This predicted ``heatmap'' represents the agent's belief with respect to the localization of the target object in question over the top-down scene floorplan. Both the question-encoder LSTM and the LingUNet models are trained end-to-end with gradients from the question-answering loss.

\xhdr{Training Details}
The visual encoder (RedNet) is first trained to perform egocentric semantic segmentation via pixel-wise CE loss on egocentric frames. Features from this pre-trained and frozen RedNet are used by the scene memory encoder to generate our episodic memory (via pixelwise CE loss on top-down semantic maps). Finally, the episodic memories from the pre-trained and frozen scene encoder are used to train the question-answering model. To do that, we use the ground truth answers (as described in Sec. \ref{sec:dataset-questions}) and train using a per-pixel binary cross entropy loss that encourages the model to correctly classify each "cell” in the top-down map as belonging to the answer category or not. Since the optimization process in our case is dealing with a severe class imbalance problem (only a few hundreds of pixels belong to the answer class among several tens of thousands of ``background'' pixels), we leverage the dynamic weighting properties of Focal loss \cite{focal} that adjusts the contribution of the easily classified ``background'' pixels to the overall loss in order to stabilize training for our model. In our experiments, we also found that setting the bias of the last layer to the ratio of the number of positive to negative samples and using the normalization trick from \cite{focal} helps.

%% file: sections/5_baselines.tex
\section{Baselines}
\label{sec:baselines}
In this section, we present details of a range of competitive baselines that we compare our approach against.

\xhdr{Language-only (\textcolor{blue}{LangOnly})}. 
We evaluate baselines which answer questions from the language input alone for EMQA. Such baselines have been shown to demonstrate competitive performance for embodied question answering tasks \cite{das2017embodied, wijmans2019embodied}. Specifically, we drop the episodic scene memory features (while keeping the temporal features) from our inputs and train the question-answering model to predict the answer to the input question. Performance of this baseline is an indication of the spatial biases present in the dataset (are beds almost always present in the same corner of the map?).

\xhdr{Egocentric semantic segmentation (\textcolor{blue}{EgoSemSeg})}. This baseline serves as a naive, ``off-the-shelf'' solution for the EMQA task. We perform semantic segmentation on each of the egocentric RGB frames that comprise the scene tour (using the same pre-trained RedNet model as used in our approach).
Then, we extract the subset of the model's predictions corresponding to the object in question and that serves as the final prediction for this baseline. 

\xhdr{Decoding top-down semantic labels (\textcolor{blue}{SMNetDecoder})}. In this baseline, we use the decoding section of the network that was used to pre-train our scene memory features and directly predict the top-down semantic segmentation of the floorplan as seen during the tour. We follow that with the same step as above: extracting the subset of top-down prediction pixels corresponding to the object in question to get the answer prediction for this baseline. Note that both the EgoSemSeg and SMNetDecoder baselines do not have access to the temporal features.

\begin{figure*}
    \centering
    \includegraphics[clip, trim=0cm 0cm 0cm 0cm, width=1.00\textwidth]{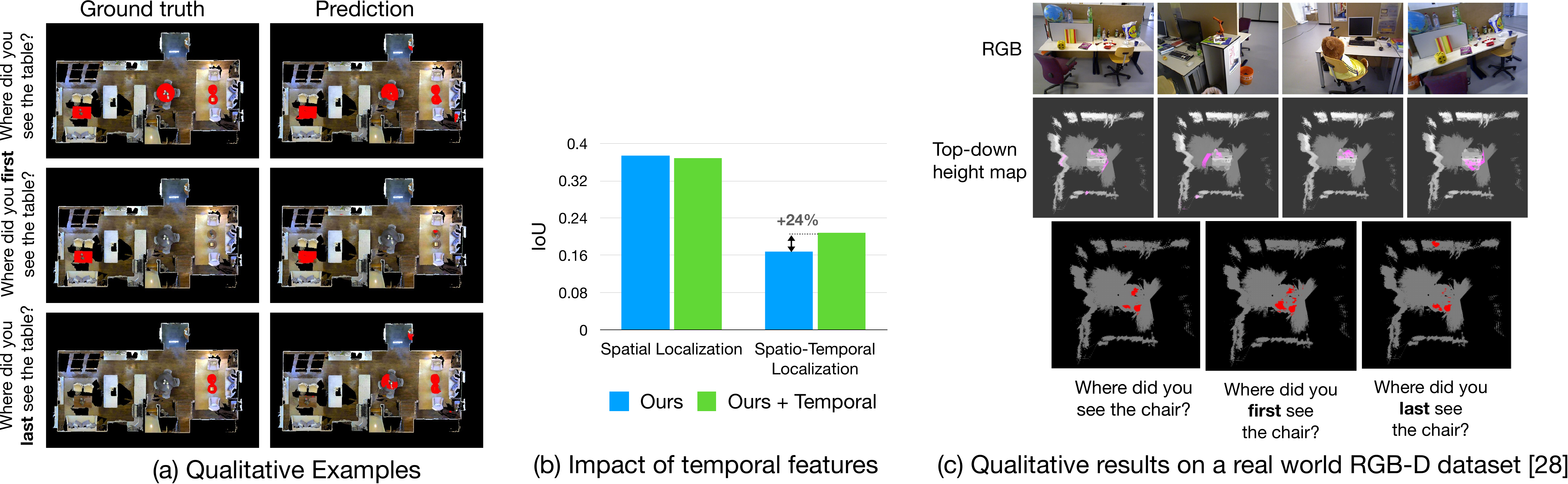}
    \vspace{0.05in}
    \caption{(a) Qualitative examples of the model's outputs in the top-down map output space from the test split of the Matterport \cite{Matterport3D} dataset. (b) The improvement in performance due to temporal features. (c) Qualitative results of zero-shot generalization of our model to real-world RGB-D dataset \cite{sturm2012benchmark}.}
    \label{fig:qual-examples-temporal-feats-compare}
\end{figure*}

\xhdr{Buffer of egocentric features as scene memory}. For this family of baselines, we store a buffer of visual features extracted from the egocentric RGB-D frames of the tour. These features are extracted via the same pre-trained RedNet model used in our approach as well as the EgoSemSeg and SMNetDecoder baselines. We then condense the per-step features in the buffer using the following different techniques to give rise to specific instantiations of baselines from prior work: (a) averaging \cite{eslami2018neural} (\textbf{\textcolor{blue}{EgoBuffer-Avg}}), (b) GRU \cite{bakker2001reinforcement} (\textbf{\textcolor{blue}{EgoBuffer-GRU}}) and (c) question-conditioned, scaled, dot-product attention \cite{fang2019scene} (\textbf{\textcolor{blue}{EgoBuffer-Attn}}).

Having generated the 1-D scene embedding vector using either of the above approaches, we use a network of de-convolutional layers to generate a top-down, 2D ``heatmap'' prediction of the agent's answer. For more details about the architecture of these baselines, please refer to the Supplementary material. Note that these baselines implicitly convert the scene embedding derived from egocentric observations into an allocentric top-down answer map (via the de-convolutional layers). On the contrary, our model has this transformation explicitly baked-in via geometrically consistent projections of egocentric features.

%% file: sections/6_results.tex
\section{Experimental Results}
\label{sec:results}

\textbf{Metrics}.
Our models generate a binary segmentation map (``answer'' v/s ``background'' pixels) as their output. Therefore, we report from the suite of segmentation metrics for evaluating the output answer localization. Specifically, for each data point (tour+question+answer), we compute: the precision, recall and the intersection-over-union (IoU) between the predicted and GT binary answer maps for the question.
We report the aforementioned metrics, averaged across the test splits of our dataset of tours.

\xhdr{Quantitative Results}. We report results for our model's predictions in both the top-down map as well as egocentric tour output modalities in Tab. \ref{tab:results-topdown}. As stated in Sec. \ref{sec:intro}, a grounding of answers into the topdown floorplan is equivalent to a localization within the egocentric pixels of the agent tour -- we simply back-project top-down map pixel predictions onto the agent's egocentric frame of reference. Therefore, for simplicity, we discuss trends in the top-down map space in the subsequent text.

We outperform the SMNetDecoder \cite{cartillier2020} baseline with $8.2$\%, $42$\% gains in IoU, recall respectively. This is due to the combination of two factors: the SMNetDecoder baseline doesn't encode knowledge about the temporal information of tours and our proposed model offers a better mechanism to ground the semantics of questions into the top-down map features via the more expressive LingUNet-based question-answering model. To isolate the gains due to the availability of temporal features, we also train a variant of our model without the same. We see that, even in the absence of temporal features, we are able to out-perform SMNetDecoder (recall of $60.81$ v/s $43.86$, IoU of $27.42$ v/s $26.92$). Moreover, the EgoSemSeg baseline performs worse than SMNetDecoder baseline (and by association, our model). This empirically demonstrates the superiority of our proposed approach over naive, ``off-the-shelf'' solutions for the task. This is consistent with observations made by prior work \cite{cartillier2020}.

All the baselines that rely on a buffer of egocentric RGB-D frame features as scene memories fail spectacularly at the task. This further verifies the hypothesis that compressed, 1-D representations of scenes are woefully inadequate for tasks such as ours. Encoding spatial knowledge of how objects are laid out in scenes and temporal information regarding when they were observed during a tour into such representations and then decoding precise localizations of answers from such scene memories is an extremely challenging problem. Our findings invalidate such memory representations from being used for our task.

In Fig. \ref{fig:qual-examples-temporal-feats-compare}, we also show qualitatively  that our agent learns to distinguish all, first- and last-seen instances of tables within a given scene (refer to Suppl. for more qualitative examples).

\begin{figure*}
    \centering
    \includegraphics[clip,trim=0cm 0cm 0cm 0cm,width=1.0\textwidth]{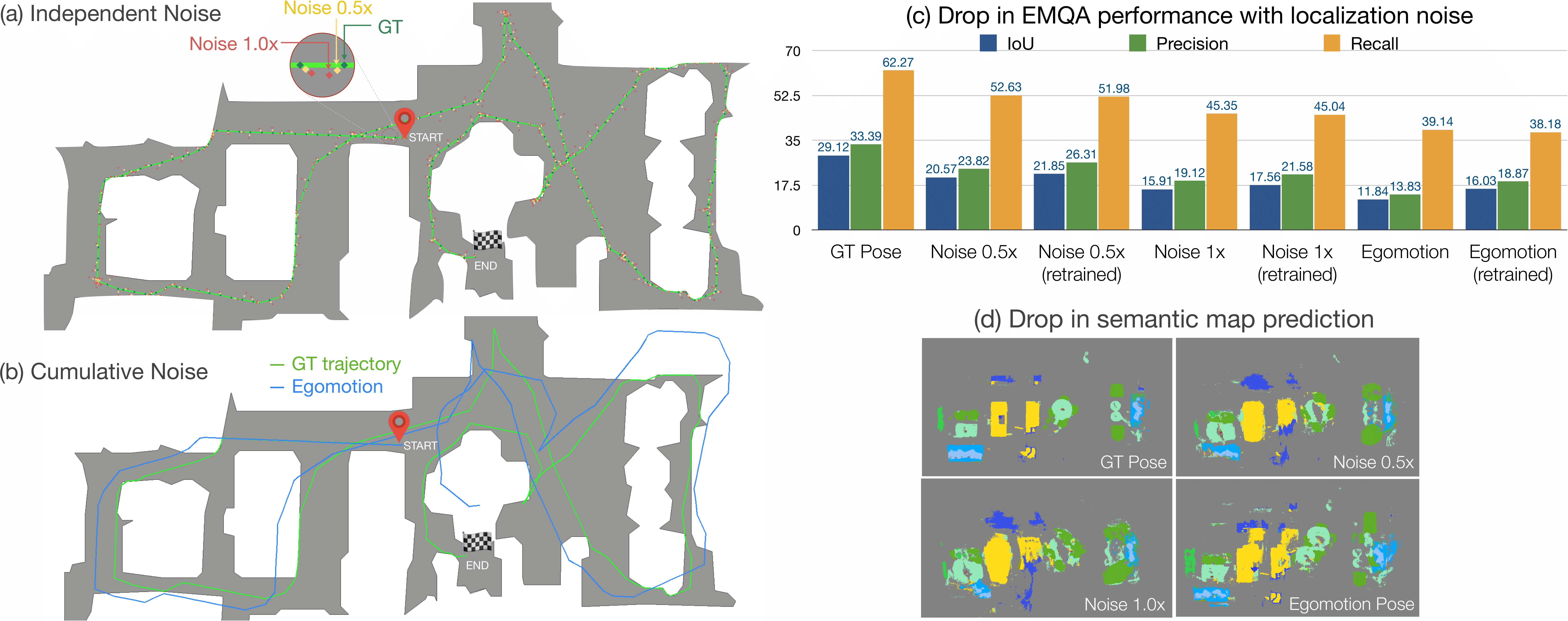}
    \vspace{0.05in}
    \caption{Samples of noisy tours obtained by adding (a) noise sampled from LoCoBot \cite{murali2019pyrobot} and (b) using a visual odometry model \cite{zhao2021surprising}. (c) Quantitative evaluation of our models under these noisy pose settings (d) Qualitative examples of noisy semantic map predictions from the Matterport \cite{Matterport3D} dataset.}
    \label{fig:noise}
\end{figure*}

\xhdr{Temporal features help}. As stated in Sec. \ref{sec:models},  having knowledge of when objects were observed during the tour is critical for answering temporal localization questions. To further elucidate this claim, we break down the performance of both variants of our model (Ours v/s Ours(+temporal) in Tab. \ref{tab:results-topdown}) by question types (spatial and spatio-temporal). As shown in Fig. \ref{fig:qual-examples-temporal-feats-compare} (b), we see a $24\%$ relative improvement in IoU for spatio-temporal localization questions upon the addition of temporal features. There is no significant impact on metrics for spatial questions.

\xhdr{Sim2Real Robustness}.
Moving beyond simulation, we analyze the robustness of our models to typical sources of noise that might arise from the real-world deployment of such systems. First, we test our EMQA model trained in simulation \cite{Matterport3D} on raw video sequences captured in the real world. We use the RGB-D observations + camera poses from the RGB-D SLAM benchmark in \cite{sturm2012benchmark} which presents a significantly challenging test-bed with high-fidelity conditions for egocentric AR applications: noisy depth+pose and head-worn camera jitter. Despite these challenges, our approach provides promising results with zero-shot generalization (\ref{fig:qual-examples-temporal-feats-compare} (c)). Without any fine-tuning, the agent is able to ground answers to questions and reasonably distinguish between first and last seen instances of objects.

Second, following prior work \cite{datta2020integrating,zhao2021surprising}, we remove the assumption of oracle indoor localization and investigate our models under conditions of noisy pose. Specifically, we perturb the ground-truth pose sequence from our dataset in two ways. First, we add noise (independently, at every step) from a distribution estimated with samples collected by a LoCoBot \cite{murali2019pyrobot} (Fig. \ref{fig:noise} (a)). Second, we predict the relative pose change between two successive steps via a state-of-the-art visual odometry model \cite{zhao2021surprising} and integrate these estimates over the trajectory to maintain a noisy estimate of the current pose (Fig. \ref{fig:noise} (b)). The latter is more realistic as it takes into account drift in the agent's pose estimates due to cascading errors along the trajectory.

As expected, when EMQA models trained with oracle localization inputs are evaluated with noisy pose, the quality of the scene representations (Fig. \ref{fig:noise} (d)) and the task metrics (Fig. \ref{fig:noise} (c)) drops proportional to the severity and nature (independent v/s cumulative) of noise being added. We find that the IoU of our proposed model drops by 29\%, as compared to a 36\% drop for our best-performing baseline, indicating that our model is more resilient to the added noise. Finally, we also show that re-training our models (both the SMNet scene encoder and LingUNet question-answering module) in the noisy settings allows us to regain some of this lost performance (increase in IoU and precision across all three noise models) in Fig. \ref{fig:noise} (c).
Refer to Suppl. for more details.

\xhdr{Limitations}.
Our approach involves building static scene maps which restricts the setup to questions about objects (such as furniture items) whose positions in the scene remain largely fixed. One way to overcome this is to update the scene maps (via re-sampling the agent tours) at sufficient frequency so that the constructed scene maps more closely approximate the current environment state.

%% file: sections/7_conclusion.tex
\section{Conclusion}
We study the task of question-answering in 3D environments with the goal of egocentric personal AI assistants. Towards that end, we propose a model that builds semantic feature representations of scenes as its episodic memory. We show that exploiting such bottleneck scene representations can enable the agent to effectively answer questions about the scene and demonstrate its superiority over strong baselines. Our investigations into the robustness of such a system to different forms of noise in its inputs present promising evidence for future research towards deploying such agents in egocentric AR devices for the real-world.

\vspace{0.5em}

\xhdr{Acknowledgements}: The Georgia Tech effort was supported in part by NSF, ONR YIP, and ARO PECASE. The views and conclusions contained herein are those of the authors and should not be interpreted as necessarily representing the official policies or endorsements, either expressed or implied, of the U.S. Government, or any sponsor.

%% file: sections/appendix.tex
\section{Additional Dataset Details}
The Matterport3D \cite{Matterport3D} meshes are annotated with 40 object categories. Following prior work \cite{cartillier2020}, we work with the $12$ most commonly occurring object types, leaving out objects such as walls, curtains that are rendered as a thin array of pixels on the top-down map. The list of the 12 object categories used in our work is as follows: shelving, fireplace, bed, table, plant, drawers, counter, cabinet, cushion, sink, sofa and chair.

\begin{figure*}[h]
    \centering
    \includegraphics[clip, trim=0cm 0cm 0cm 0cm, scale=0.30]{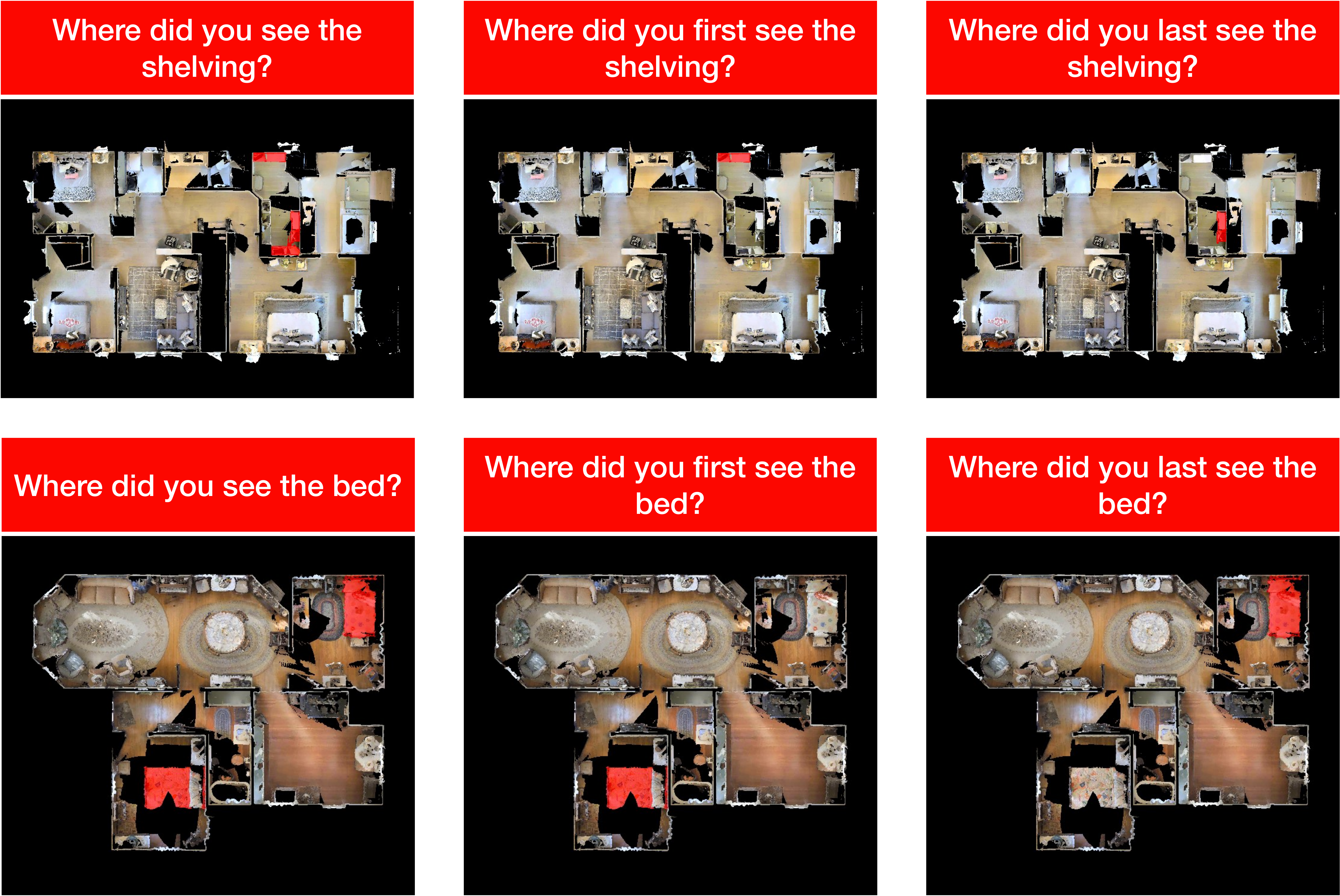}
    \vspace{0.15in}
    \caption{This figure illustrates more examples of questions grounded in Matterport \cite{Matterport3D} scenes from our proposed EMQA dataset.}
    \label{fig:dataset-qual-examples-suppl}
\end{figure*}

Fig. \ref{fig:dataset-qual-examples-suppl} shows additional qualitative examples of questions and associated ground-truth, top-down answer maps from our dataset.

\begin{figure*}[h]
    \centering
    \includegraphics[clip, trim=0cm 0cm 0cm 0cm, scale=0.40]{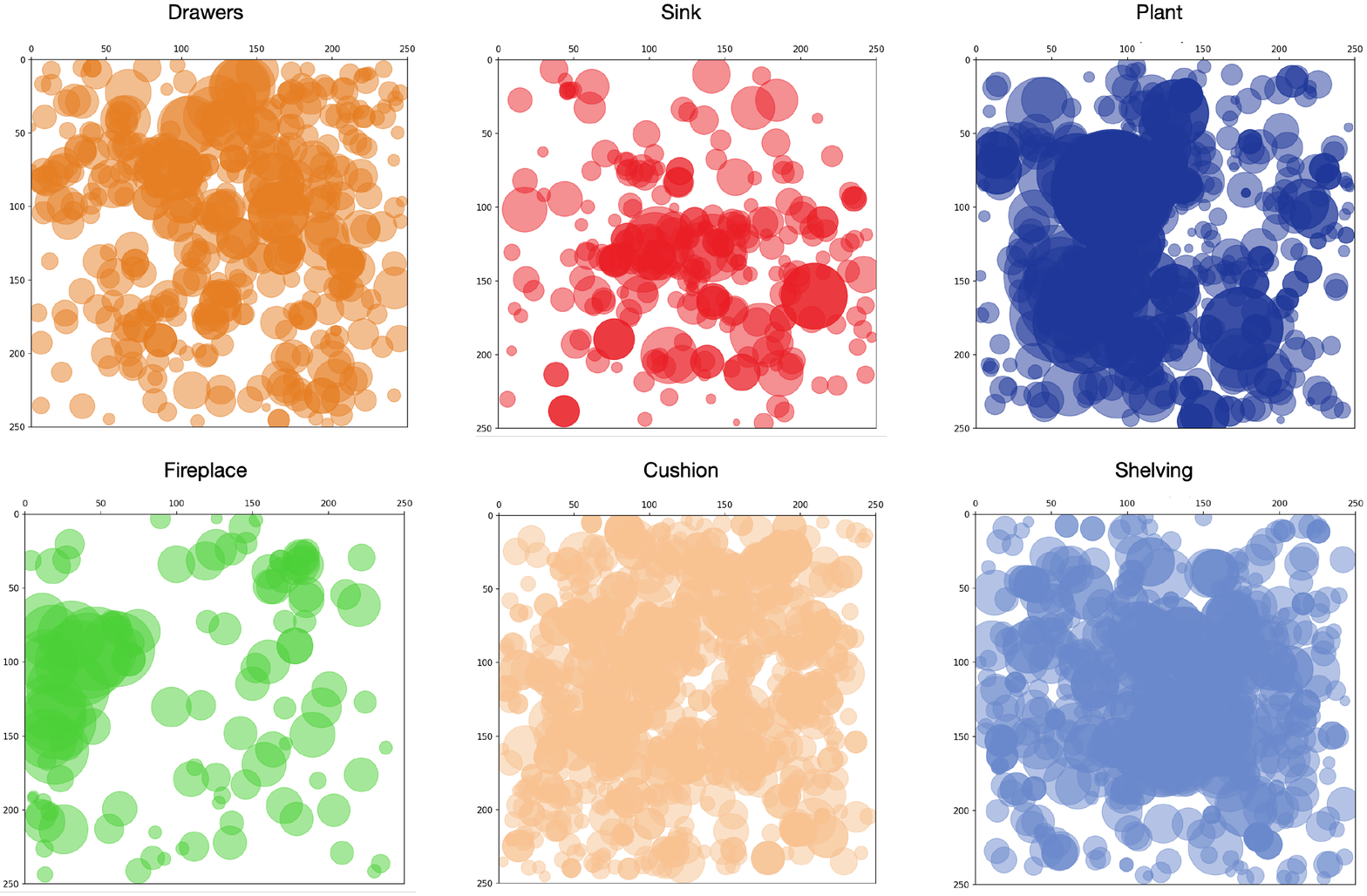}
    \caption{Distribution of spatial arrangement and size of object instances across $6$ different object categories. Each instance is represented by a circle whose center and radius correspond to the location and size of the object, as viewed on the top-down map.}
    \label{fig:dataset-stats-obj-dist}
\end{figure*}

\begin{figure*}[h]
    \centering
    \includegraphics[clip, trim=0cm 0cm 0cm 0cm, scale=0.40]{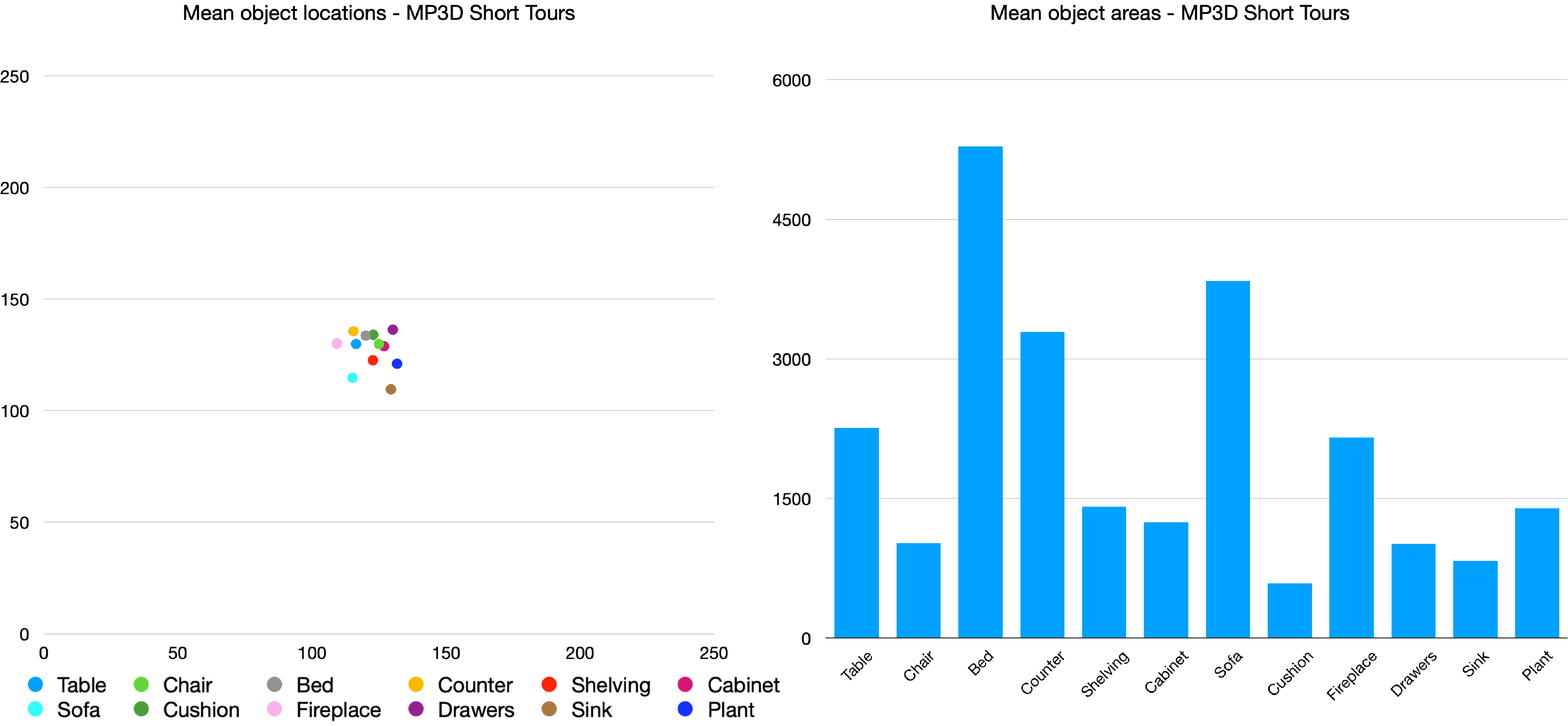}
    \caption{[Left] Scatter-plot showing the mean locations of all instances across a given object category, for all object categories. [Right] A distribution of the mean object size on the top-down map across all categories demonstrating the spectrum of object sizes in our dataset.}
    \label{fig:dataset-stats-overall-dist}
\end{figure*}

We also show additional analysis of the distribution of locations and sizes of objects from our questions (Fig. \ref{fig:dataset-stats-obj-dist} and \ref{fig:dataset-stats-overall-dist}). Fig. \ref{fig:dataset-stats-obj-dist} visualizes the spatial distribution of 6 object types (drawers, sink, plant, fireplace, cushion and shelving) over the $250 \times 250$ spatial maps across the ``short tours'' data splits (recall that this is the dataset that we train our models on). Each object instance is represented by a circle with the center of the circle corresponding to the location of an object instance and the radius representing its size (on the top-down map). The sizes of object instances are measured in terms of the number of pixels. As mentioned in Sec. 3 (main paper), all top-down maps are of a fixed $2$cm $\times 2$cm resolution, therefore the number of pixels covered by an object on the top-down map has a direct correlation with its size in the real world. 

As evident from Fig. \ref{fig:dataset-stats-obj-dist}, the objects demonstrate a good distribution over spatial locations in the map, thereby alleviating dataset biases such as ``are sinks almost always found in specific corners of the map?''. Recall that, the fact that we beat the \textcolor{blue}{LangOnly} baseline by almost $150$\% is also indicative of the fact that our models are free from such biases as well.

Additionally, Fig. \ref{fig:dataset-stats-overall-dist} (left) shows the mean object locations for all $12$ categories across the ``short tours'' data splits. Notice that, consistent with Fig. \ref{fig:dataset-stats-obj-dist}, the mean locations for nearly all object categories are close to the center of the $250 \times 250$ map. This further confirms a comprehensive distribution of objects across all spatial locations in the map. Fig. \ref{fig:dataset-stats-overall-dist} (right) shows a distribution of average object sizes per category. Here, we can see that our dataset involves asking questions about objects of varying sizes -- from typically small objects such as cushions to bigger ones such as beds.

\section{Additional Model+Training Details}
\xhdr{Details of the temporal features}.
In this section, we provide more details regarding the construction of the temporal features component of our scene memories. At a high level, we save temporal features at every (i, j) gridcell on the top-down map. To do that, we first represent each video tour (of varying lengths) as a sequence of 20 (determined heuristically) equal-length segments. For every tour segment, we then compute the subset of top-down grid-cells that were observed by the agent during that part of the tour. This induces a binary topdown map, per tour segment, indicating whether a given (i, j) was observed during that part of the tour or not. Aggregated over all segments via channel-wise stacking, we get a 20-dim multi-hot vector at each (i, j). Intuitively, these 20-bits for any top-down location (i, j) indicate the time instances (in terms of segments) during the tour when that metric location was observed.

\begin{figure*}[h]
    \centering
    \includegraphics[clip, trim=0cm 0cm 0cm 0cm, scale=0.60]{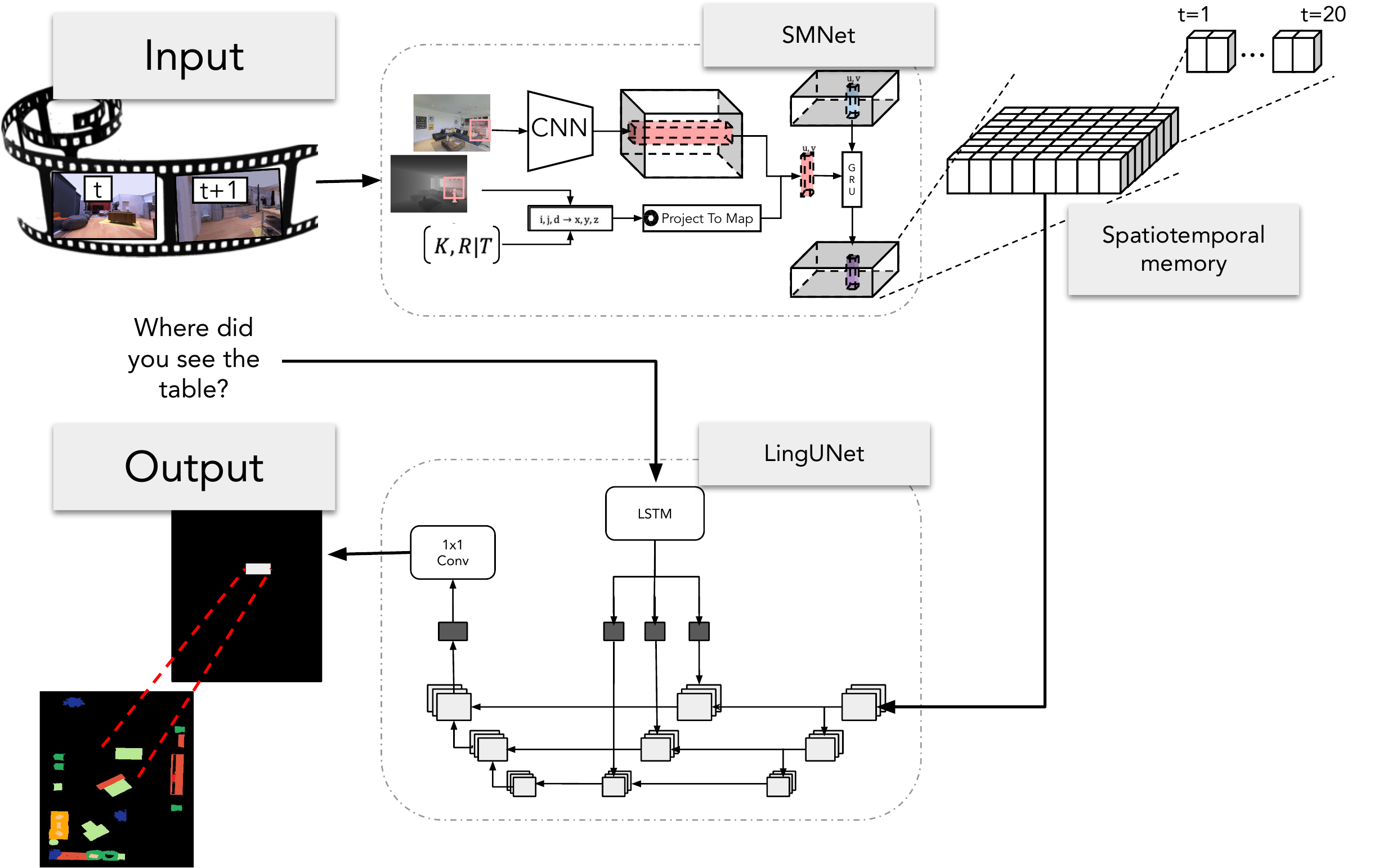}
    \vspace{0.1in}
    \caption{A schematic diagram of our proposed EMQA agent. Our agent first constructs an episodic memory representation of the tour and then grounds answers to questions on the top-down scene floorplan using a LingUNet-based answering model. A sample tour from the Matterport \cite{Matterport3D} is shown in the figure.}
    \label{fig:model}
\end{figure*}

\xhdr{Details of the LingUNet architecture}.
Fig. 3 in the main paper shows a schematic diagram of our EMQA agent. We show the same in Fig. \ref{fig:model} for the readers' convenience and provide more details about our LingUNet-based question-answering module in this section. Recall (from Sec. 4 in the main paper) that our LingUNet-based question-answering model takes the constructed scene memory features and an embedding of the question as an input and generates a top-down heatmap of the agent's answer prediction as its output. We use a $3$-layer LingUNet architecture, as shown in Fig. \ref{fig:model}. The module receives a tensor of dimensions $256 \times 250 \times 250$ (the spatio-temporal memory tensor) and a $64$-dim question LSTM embedding as inputs. The scene memory tensor is encoded through  a series of three, $3 \times 3$ Conv. layers and the question embedding is L2-normalized and used to generate three convolutional filters via FC-layers. These question-conditioned convolutional filters operate on the intermediate feature outputs from the $3$ encoder Conv. layers to generate the language-conditioned skip connections. Finally, the encoded feature volumes are passed through a series of $3$ de-convolutional layers (see Fig. \ref{fig:model}), with added skip connections to generate the final output map: a tensor containing a $128$-dim feature vector for each of the $250 \times 250$ spatial locations of the input map.

The output feature map from the LingUNet model is post-processed by a $1 \times 1$ Conv. layer, followed by Sigmoid non-linearity to generate the agent's prediction score for each of the $250 \times 250$ spatial map cell.

\xhdr{Other training hyperparameters}.
As stated in Sec. 4 (main paper), we use a dynamically-weighted variant of the binary cross entropy loss (Focal loss \cite{focal}), applied per ``pixel'' on the top-down map output to train our model to correctly predict whether each metric ``pixel'' on the map belongs to the answer category or not. We set the gamma parameter in our Focal loss implementation to $2.0$ and use the Adam optimizer with a learning rate of $2$e$-4$ and weight-decay of $4$e$-4$. To circumvent memory issues during training, our data loader returns singleton batches of questions (and associated scene memory features). However, we accumulate gradients and update the model weights every $B = 4$ iterations to simulate an effective batch-size of $4$.

\xhdr{Details about EgoBuffer baselines}.
In this section, we provide more architectural details about the \textcolor{blue}{EgoBuffer-*} baselines (Sec. 5 in the main paper). As mentioned in Sec. 5, as a first step, we compute egocentric feature volumes for every step of the agent tour via a pre-trained RedNet model (the same pre-trained RedNet model weights are used across all our models and baselines).

For the \textcolor{blue}{EgoBuffer-Avg} baseline, we simply average these per-step, convolutional feature volumes and pass them through a $4$-layer convolutional network (followed by a linear layer) to get a flattened, $1$-D vector representation of the agent tour.

For the \textcolor{blue}{EgoBuffer-GRU} baseline, we instead compute the encoded, flattened, $1$-D feature representation at every step and then pass them through a GRU. The hidden state from the final time step is the representation of the agent's tour.

And finally, for the \textcolor{blue}{EgoBuffer-Attn.} baseline, instead of sending the per-step, flattened egocentric features as inputs to a GRU, we perform a scaled, dot-product attention, conditioned on the question-embedding to obtain the tour representation.

Once we obtain the tour representation using the methods described above, we append the features with the question embedding and then decode the multi-modal (question+scene) features into a $250 \times 250$ feature map via a $5$-layer de-convolutional network. Finally, use a $1 \times 1$ Conv + Sigmoid layer to get the answer prediction scores.

Also note that, similar to our proposed model, the family of \textcolor{blue}{EgoBuffer-*} baselines are trained on fixed-size, $20$-step ``short'' tours. During evaluation on ``full tours'', we split them into consecutive chunks of $20$-steps each, generate a $250 \times 250$ output for each such ``short tour'' segment and then combine them into the full-scale environment map to get the final output (using pose information).

% short tour results (topdown/ego)
\begin{table*}[h]
    \rowcolors{2}{gray!10}{white}
    \small
    \centering
        \begin{tabular}{l c c  c c c c c c}
        \toprule
        && \multicolumn{3}{c}{\textbf{{\footnotesize Top-down map output space}}} && \multicolumn{3}{c}{\textbf{{\footnotesize Egocentric pixel output space}}}  \\
        \cmidrule{3-5} \cmidrule{7-9}
        \textbf{{\footnotesize Method}} && {\footnotesize IoU} & {\footnotesize Recall} & {\footnotesize Precision} && {\footnotesize IoU} & {\footnotesize Recall} &  {\footnotesize Precision} \\
        \cmidrule{1-9}
        LangOnly && 15.09 {\tiny $\pm$ 0.22} & 22.36 {\tiny $\pm$ 0.26} & 29.56 {\tiny $\pm$ 0.41} && 15.23 {\tiny $\pm$ 0.23} & 23.19 {\tiny $\pm$ 0.27} & 29.61 {\tiny $\pm$ 0.43} \\
        EgoSemSeg && 27.51 {\tiny $\pm$ 0.32} & 39.24 {\tiny $\pm$ 0.48} & \textbf{55.53} {\tiny $\pm$ 0.58} && 29.05 {\tiny $\pm$ 0.36} & 39.45 {\tiny $\pm$ 0.50} & \textbf{59.57} {\tiny $\pm$ 0.55} \\
        SMNetDecoder && 30.76 {\tiny $\pm$ 0.37} & 45.06 {\tiny $\pm$ 0.46} & 54.51 {\tiny $\pm$ 0.37} && 31.36 {\tiny $\pm$ 0.31} & 45.86 {\tiny $\pm$ 0.23} & 55.52 {\tiny $\pm$ 0.54} \\

        \hline
        EgoBuffer-Avg \cite{eslami2018neural} && 2.46 {\tiny $\pm$ 0.09} & 4.53 {\tiny $\pm$ 0.17} & 8.10 {\tiny $\pm$ 0.26} && 2.18 {\tiny $\pm$ 0.09} & 4.29 {\tiny $\pm$ 0.18} & 7.98 {\tiny $\pm$ 0.28} \\
        EgoBuffer-GRU \cite{bakker2001reinforcement} && 1.39 {\tiny $\pm$ 0.06} & 2.02 {\tiny $\pm$ 0.11} & 9.09 {\tiny $\pm$ 0.22} && 1.34 {\tiny $\pm$ 0.07} & 2.08 {\tiny $\pm$ 0.12} & 9.36 {\tiny $\pm$ 0.23}\\
        EgoBuffer-Attn \cite{fang2019scene} && 2.18 {\tiny $\pm$ 0.07} & 3.67 {\tiny $\pm$ 0.12} & 10.08 {\tiny $\pm$ 0.24} && 2.01 {\tiny $\pm$ 0.07} & 3.66 {\tiny $\pm$ 0.12} & 10.15 {\tiny $\pm$ 0.26} \\
        \hline
        
        \textbf{Ours} && 36.09 {\tiny $\pm$ 0.53} & 48.03 {\tiny $\pm$ 0.62} & 53.04 {\tiny $\pm$ 0.51} && 36.66 {\tiny $\pm$ 0.27} & 49.64 {\tiny $\pm$ 0.41} & 53.12 {\tiny $\pm$ 0.32}\\
        \textbf{Ours} (+temporal) && \textbf{37.72} {\tiny $\pm$ 0.45} & \textbf{49.88} {\tiny $\pm$ 0.55} & 53.48 {\tiny $\pm$ 0.64} && \textbf{38.29} {\tiny $\pm$ 0.44} & \textbf{51.47} {\tiny $\pm$ 0.58} & 53.48 {\tiny $\pm$ 0.59} \\
        \bottomrule
        \end{tabular}
    \\ [8pt]
    \caption{EMQA results on ``short tours'' for our proposed model and baselines in the ``top-down map'' and ``egocentric pixel'' output space.}
    \label{tab:results-short-tours}
\end{table*}

\section{Additional Quantitative EMQA Results}
As a means to circumvent memory constraints during training and as a data augmentation strategy, we subsample 20-step ``short'' tours from the full-scale exploration tours in our dataset (see Sec. 3, sub-heading, ``Generating ``short'' tours for training'' in the main paper for more details). During training, our model learns to build top-down maps (of smaller spatial dimensions: $250 \times 250$ than that of the full scene) and localize answers to questions on the map from these 20-step tours. During evaluation, we test its generalization to building full-scale maps from the entire exploration trajectory (and not just 20-step tours). We presented results (both quantitative as well as qualitative) for this full-scale generalization in the main paper. In this section, for completeness, we also provide results of all our models and baselines on the 20-step ``short tours'' in Tab. \ref{tab:results-short-tours}. We also show qualitative examples of predictions made by our agent for these 20-step short tours in Fig. \ref{fig:qual-examples-suppl-short}. All trends observed and discussed in the main paper (Section 6) hold true for Tab. \ref{tab:results-short-tours} as well.

We’d like to reiterate (from Section 3 in the main paper) that generating and using these ``short tours'' merely happens during train. The focus of our task (as well as the subject of all our evaluations/analysis in the main paper) is on the full-scale exploration tours and maps for entire scenes.

\section{Additional Qualitative EMQA Results}
\begin{figure*}[h]
    \centering
    \includegraphics[clip, trim=0cm 0cm 0cm 0cm, width=1.00\textwidth]{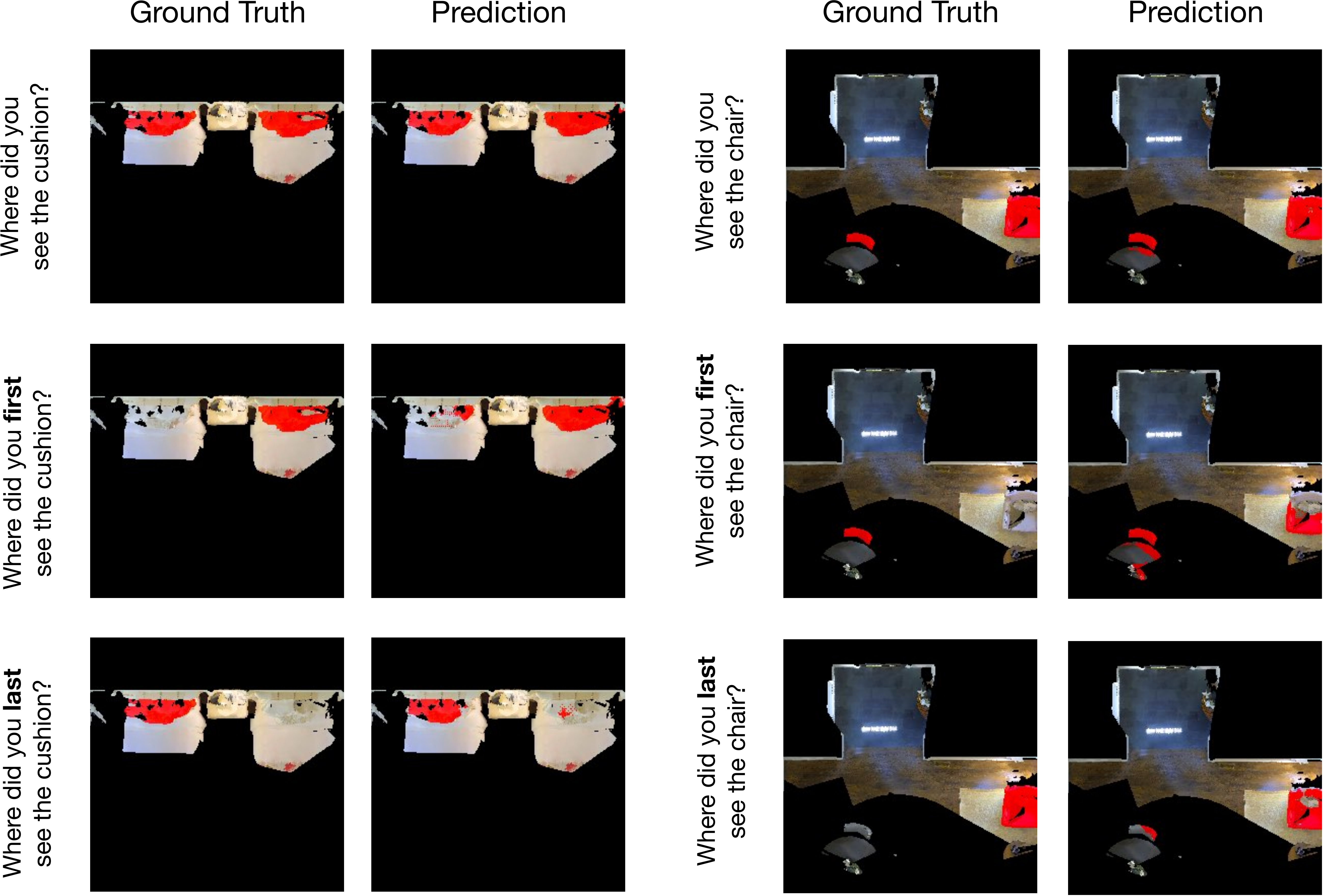}
    \vspace{0.1in}
    \caption{We provide qualitative results of our proposed EMQA agents grounding answers to questions onto the top-down environment floorplan for ``short'' tours in Matterport \cite{Matterport3D}.}
    \label{fig:qual-examples-suppl-short}
\end{figure*}

\begin{figure*}[h]
    \centering
    \includegraphics[clip, trim=0cm 0cm 0cm 0cm, width=1.00\textwidth]{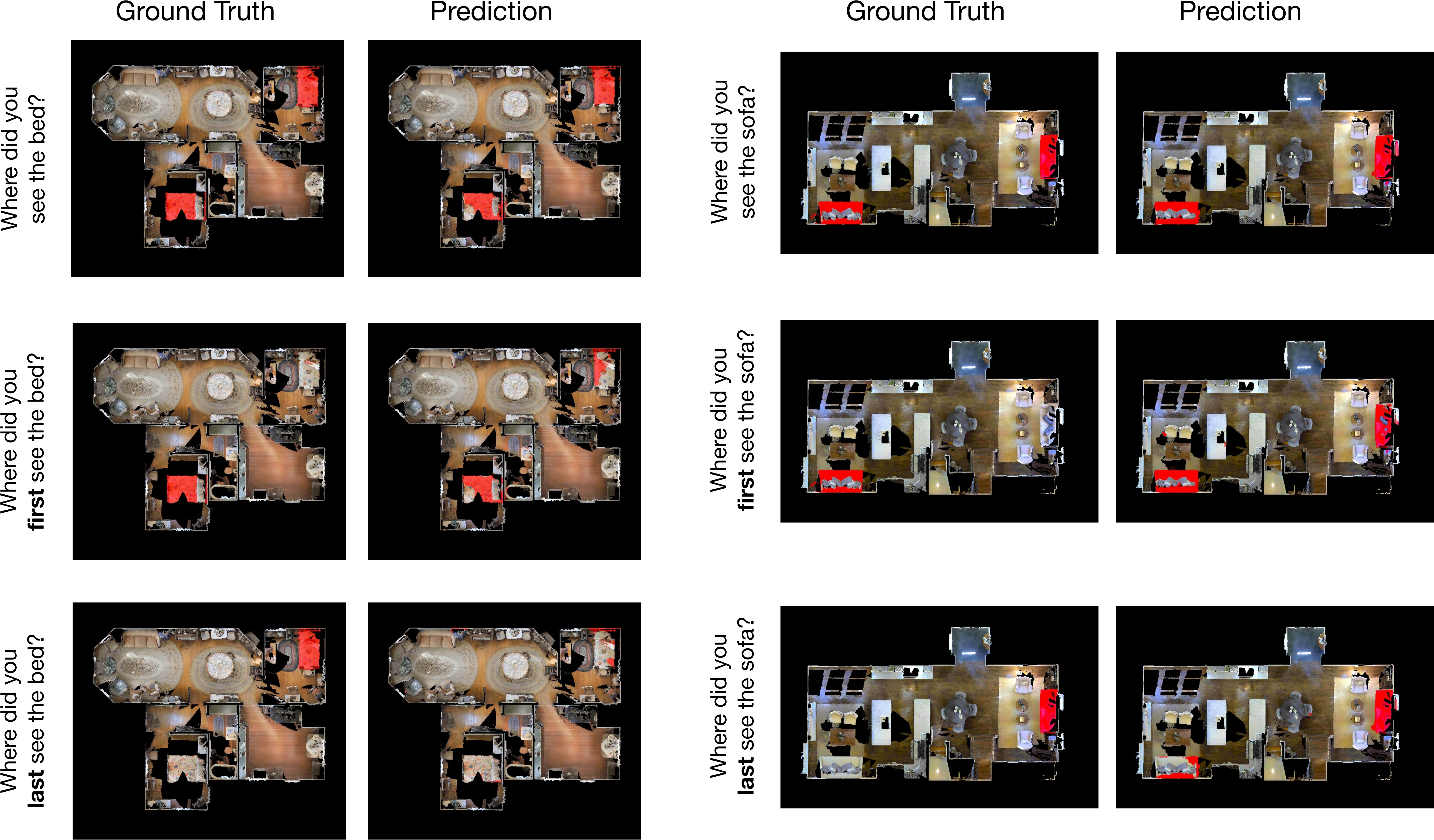}
    \vspace{0.1in}
    \caption{We provide additional qualitative results of our proposed EMQA agents grounding answers to questions onto the top-down environment floorplan for ``full'' tours in Matterport \cite{Matterport3D}.}
    \label{fig:qual-examples-suppl-full}
\end{figure*}

We show additional qualitative results on both ``short'' (Fig. \ref{fig:qual-examples-suppl-short}) and ``full'' (Fig. \ref{fig:qual-examples-suppl-full}) tours for our proposed EMQA model.

\section{Sim2Real: Real-world RGBD Results}
Recall that Fig. 4 (c) in the main paper had qualitative results of evaluating our model on a more challenging, real-world RGB-D dataset \cite{sturm2012benchmark} with imperfect depth+pose and camera jitter. Here, we provide some additional details about the same.

We perform zero-shot evaluation tests on this dataset -- we construct scene memories using our pre-trained SMNet \cite{cartillier2020} model, hand-craft questions relevant to the scene and generate prediction output using our pre-trained LingUNet-based question-answering module (no component of our model was fine-tuned on \cite{sturm2012benchmark}).
We pre-processed the input video by subsampling the original frame sequence (selecting every $10$th frame). We discarded visually blurry images by looking at the variance of the Laplcaian in the RGB image. Any image with a variance below a threshold of $100$ is discarded. Data inputs are grouped into tuples of (RGB, Depth and Pose) using timestamps. We allow for a max timestamp difference of at most 0.02s between modalities. The depth frame is further pre-processed through a binary erosion step with a circular element of radius 20 pixels.

In the answer to the question 'where did you first see the chair?’ (Fig. 4 (c) in the main paper), we see the first two chairs (at the bottom) of the map being detected. The third chair on the other side of the desk is also detected. Although it is not, strictly speaking, the first instance of a chair seen in the video, that chair appears at the beginning of the video. 
In the answer to the question 'where did you last see the chair', that third chair at the top of the map is not detected this time. The bottom two chairs are detected again because there are also seen last in the video.
In the answer to the question ‘where did you see the chair’, three chairs are detected. The fourth one with the stuffed animal on it has been missed.

\section{Sim2Real: Noisy Pose Experiments}
In this section, we discuss the impact of localization noise on the construction of scene memory representations and downstream question answering performance.

\subsection{Noise Models}
As discussed in Sec. 6 of the main paper (subsec: Sim2Real Robustness), we investigate the impact of two qualitatively different types of noise -- (1) noise sampled from a real-world robot \cite{murali2019pyrobot}, \textbf{independently} added to the pose at each step along the ground truth trajectory, and (2) \textbf{cumulatively} integrating per-step noisy pose change estimates derived from a visual odometry-based egomotion estimation module.

In this section, we discuss how we implement and integrate these two noise models into the pose at each step of our guided exploration tours.

\xhdr{Independent Noise}. Here, we leverage the actuation noise models derived from the real-world benchmarking \cite{kadian2020sim2real} of a physical LoCoBot \cite{murali2019pyrobot}. Specifically, this comprises the linear and rotational actuation noise models. The linear noise distribution is a bi-variate Gaussian (with a diagonal covariance matrix) that models localization inaccuracies along the X (orthogonal to the heading on the ground plane) and Z axis (direction of heading on the ground plane), whereas the rotational noise distribution is a uni-variate Gaussian for the agent's heading. We integrate these noise models in the guided tours by independently adding noise to each ground truth pose of the assistant along the tour.
Additionally, we implement this setup with two levels of noise multipliers -- 0.5x and 1x to simulate varying intensities of noise being added. An example of a trajectory with this noise model has been visualized in Fig. 5(a) in the main paper.

\xhdr{Cumulative Noise}. Beyond adding noise to existing ground truth trajectory pose, we additionally investigate a setting where our model is given an initial pose and from thereon, it estimates per-step changes in its pose solely from observations (visual odometry) and integrates these predictions along the trajectory, thereby maintaining an up-to-date (though noisy) estimate of its current pose

In contrast to the independent addition of noise to the ground truth pose at each tour step, this is a more challenging setting where the assistant maintains and works with a pose estimate completely on its own. As a result, noise picked up at each step (e.g. during collisions) cascades and accumulates throughout the rest of the trajectory. We present an example of a noisy egomotion trajectory in Fig. 5(b). We now describe the visual odometry model that we adapt from \cite{zhao2021surprising} and use to estimate per-step pose changes.

\subsection{Visual Odometry Models}
\xhdr{Model Architecture}.
The visual odometry (VO) model takes as input a pair of consecutive RGB-D observations ($o_t$, $o_{t+1}$) and estimates three relative pose transformation components -- the translation along X and Z axis, and the rotation angle ($\theta$).
For our experiments, we employ a stripped-down (bare-bones) version of the architecture proposed in \cite{zhao2021surprising}. The architecture includes a Resnet-18 \cite{he2015resnet} feature extractor backbone followed by two fully-connected (FC) layers with Dropout. We do away with the soft top-down projection and discretized depth inputs, and only use RGB+Depth observation pairs from successive steps. Additionally, as suggested by the authors \cite{zhao2021surprising}, we train action-specific models to deal with the differences in the action distributions.

\begin{figure}[h]
    \centering
    \includegraphics[clip, trim=0cm 0cm 0cm 0cm, width=0.496\textwidth]{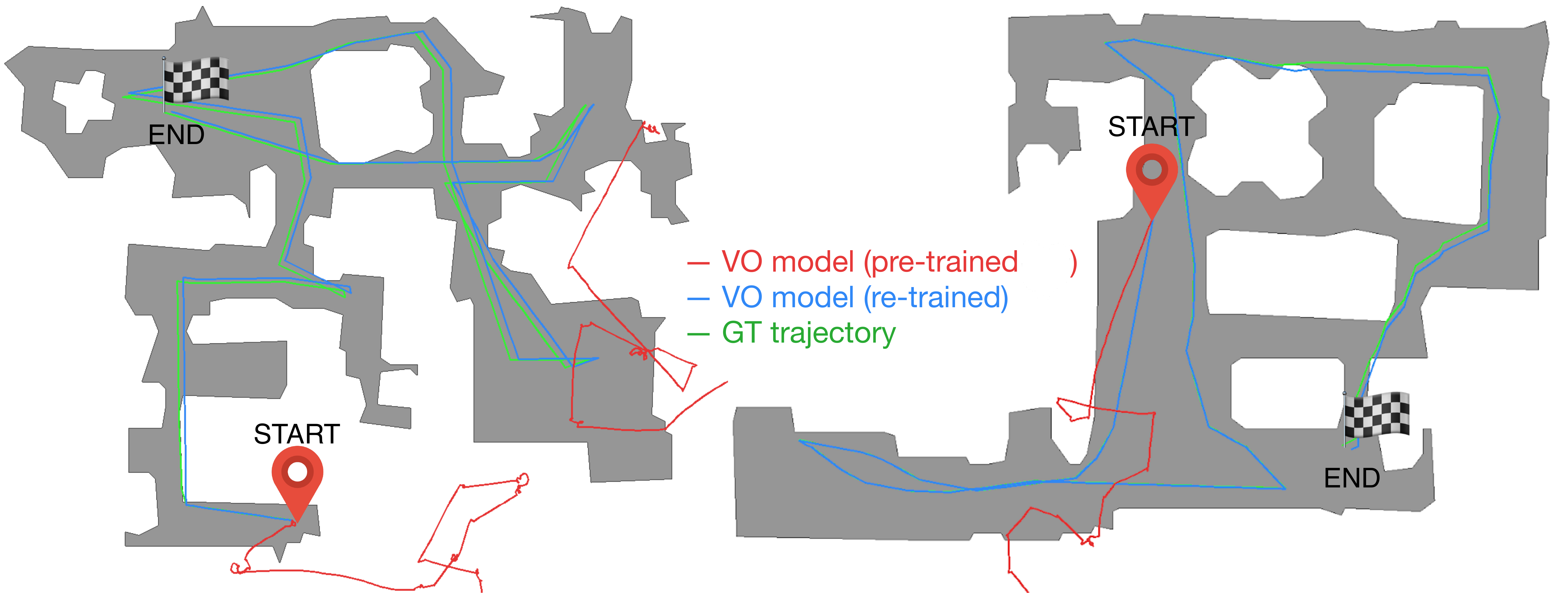}
    % \vspace{0.1in}
    \caption{Difference in egomotion predictions between the pre-trained VO model provided by the authors of \cite{zhao2021surprising} and the model we retrained on a custom dataset of 100k data points from Matterport3D \cite{Matterport3D} scenes. Apart from the difference in the scene datasets, the provided pre-trained models were trained under significantly dissimilar actuation specifications. This discrepancy called for a re-training of the VO model under the guided tour setup we use \cite{cartillier2020}.}
    \label{fig:retrained-vs-pretrained-vo}
\end{figure}

\xhdr{Dataset Preparation}.
The pre-trained VO models provided by the authors of \cite{zhao2021surprising} were trained on the Gibson dataset of 3D scans \cite{xia2018gibson}. In contrast, we work with scenes from the Matterport3D \cite{Matterport3D} dataset. More importantly, these models were trained in a setup that used agent actuation specifications (forward step: 25cm, rotation angle: 30\textdegree) that are significantly different from those used in our exploration tours (forward step: 10cm, rotation angle: 9\textdegree). Therefore, a zero-shot transfer of models pre-trained by the authors in \cite{zhao2021surprising} to our setup is likely to not work well. We qualitatively verify this in Fig. \ref{fig:retrained-vs-pretrained-vo}. Note that the trajectory formed by integrating predictions from a pre-trained VO model applied directly to our setup doesn't match the original trajectory at all.

To remedy this, we retrain the VO models from \cite{zhao2021surprising} on scenes from the Matterport3D dataset and under the actuation specificns of EMQA. To do that, we first create a separate VO dataset in Matterport3D \cite{Matterport3D}. For generating the dataset, we follow the protocol laid out in \cite{zhao2021surprising}. We sample (uniformly, at random) two navigable points in the scene and compute the shortest navigable path between the two points. Then, we sample (again, uniformly, ar random) consecutive observation pairs along these shortest-path trajectories. Using this method, we create a dataset of 100k and 20k training and validation data points.

\xhdr{Training and Evaluation}.
Following the training regime proposed by \cite{zhao2021surprising}, we train our model by jointly minimizing the regression and geometric invariance losses from \cite{zhao2021surprising}. Fig. \ref{fig:retrained-vs-pretrained-vo} \textcolor{blue}{(blue)} shows the trajectory generated via integration of predictions from the retrained VO model. We note a significant qualitative improvement in the trajectory as it much closely resembles the ground truth trajectory. In addition to that, in Tab. \ref{tab:noisy-trajectory-deviation}, we also quantitatively analyze the improvements gained through retraining by comparing the average deviation in the trajectories (as measured by the average over per-step pose RMSEs) predicted via the pre-trained and the re-trained VO models. We note that with a retrained VO model, we significantly improve the quality of our trajectories by getting $\sim6$x less deviation with the ground truth trajectory.

For completeness, Tab. \ref{tab:noisy-trajectory-deviation} also contains metrics for trajectories generated by independently adding noise to the ground-truth pose. It is evident that both (a) increasing the intensity of independently added noise and (b) moving from independent to VO-based cumulative noise leads to larger deviations in the predicted trajectories from the ground truth ($\sim$1.3x and $\sim$15x increase respectively).

\begin{table}[h]
    \rowcolors{2}{gray!10}{white}
    \small
    \centering
        \begin{tabular}{l c c c}
        \toprule
        \textbf{Method} & {RMSE (X axis)} & {RMSE (Z axis)} \\
        \cmidrule{1-3}
        Noise 0.5x (independent) & 0.068 & 0.066 \\
        Noise 1x (independent) & 0.09 & 0.088 \\
        Egomotion (re-trained) & 1.448 & 1.346 \\
        Egomotion (pre-trained \cite{zhao2021surprising}) & 11.074 & 10.386 \\
        \bottomrule
        \end{tabular}
    \\ [8pt]
    \caption{Average absolute and relative differences between the ground truth poses and their corresponding noisy estimates at each step. Here, we highlight the improvement in egomotion estimation by retraining the VO model for our setup. We also note how, due to its cumulative nature (accumulating noise along the trajectory), the error metrics are much higher for the egomotion predictions compared to the independent noise models (where noise is independently added to the pose at each step).}
    \label{tab:noisy-trajectory-deviation}
\end{table}

\begin{figure*}[h]
    \centering
    \includegraphics[clip, trim=0cm 0cm 0cm 0cm, width=1\textwidth]{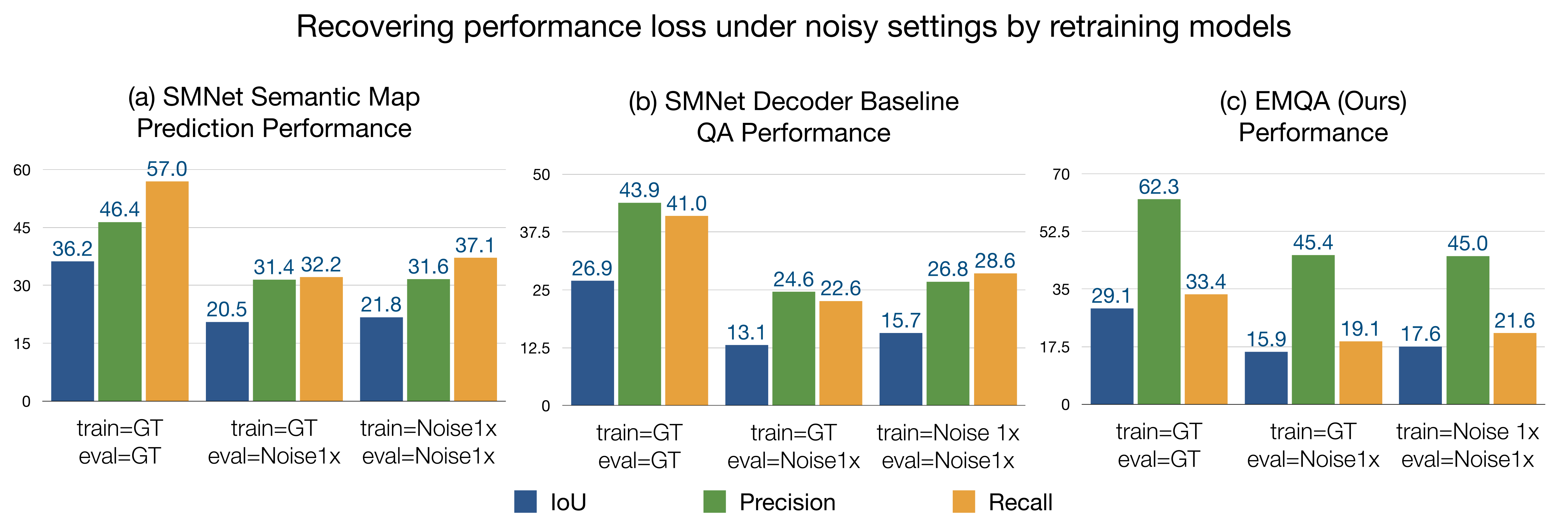}
    % \vspace{0.1in}
    \\ [8pt]
    \caption{Recovering performance loss under noisy settings in semantic map prediction (a) and question answering (b,c) by retraining SMNet and LingUNet on noisy pose data. In each sub-figure, we observe a dip in the evaluation metrics of the ground truth (GT) pose-trained model upon adding noise to the input pose. This drop is partially recovered from, by retraining the model(s) on noisy data.}
    \label{fig:noise-retraining}
\end{figure*}

\begin{figure*}[h]
    \centering
    \includegraphics[clip, trim=0cm 0cm 0cm 0cm, width=0.75\textwidth]{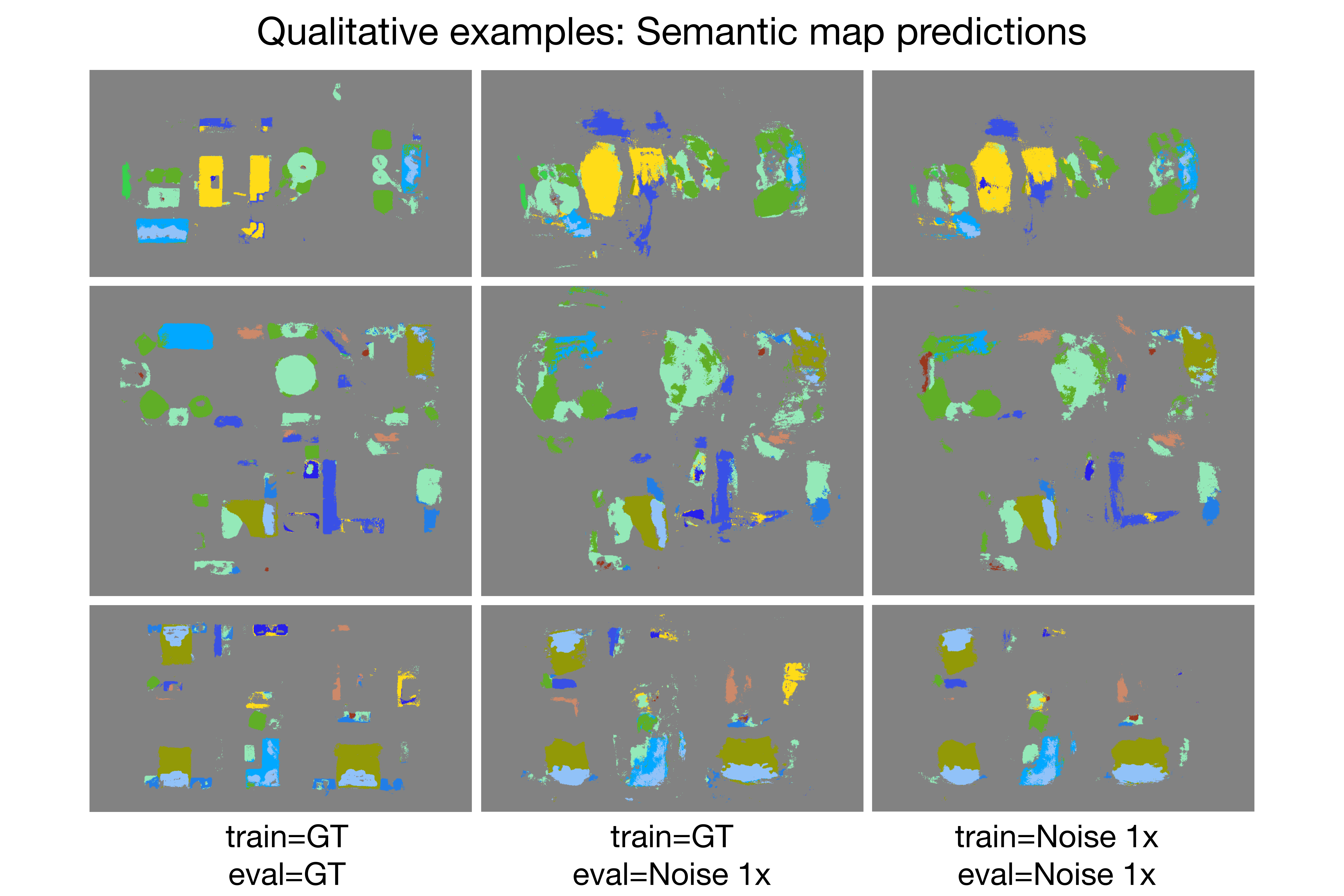}
    % \vspace{0.1in}
    \\ [8pt]
    \caption{We show qualitative examples of semantic map predictions for the three experiments described in Sec. \ref{sec:emqa_noise_eval} -- (1) our model trained and evaluated using ground truth pose (left), (2) our model trained using ground-truth pose, but evaluated in the noisy pose setting (middle) and (3) our model retrained using noisy pose (right). We see sharper boundaries and less label splatter upon retraining with noise.}
    \label{fig:qual-retraining}
\end{figure*}

\subsection{Evaluating EMQA Models on Noisy Pose Data}
\label{sec:emqa_noise_eval}
In the discussions so far, we talked about adding noise (of various types and intensities) to the pose information in our dataset. In this section, we first investigate the performance of our models under the noisy settings and then, make a case for re-training our EMQA models so that they learn to adapt to noisy pose inputs.

Towards that end, we plot the improvement in performance upon retraining across three setups in Fig. \ref{fig:noise-retraining}. In each of the sub-figures, we first plot the original model's performance (trained and evaluated using ground truth pose i.e. train=GT, eval=GT). This is followed by taking the same model (trained using ground truth  pose) and evaluating it under the noisy pose setting (i.e. train=GT, eval=Noise1x). And finally, we re-train our model so that it adapts to the noise and then evaluate this re-trained model in the noisy setting (train=Noise1x, eval=Noise1x).

We now compare the above settings across three experiments.

\xhdr{1. Semantic map prediction performance of SMNet (Fig. \ref{fig:noise-retraining} (a))}: Our method utilizes scene memory representations learnt by SMNet for downstream question answering. As a result, comparing the semantic map predictions serves as a proxy for the quality of scene representations. Here, we observe a 43\% drop in IoU on adding noise and evaluating using the original model. Upon retraining SMNet with noisy data, we see a 6\% gain on the same metric.

\xhdr{2. Question answering (QA) performance of SMNet Decoder baseline \cite{cartillier2020} (Fig. \ref{fig:noise-retraining} (b))}:
Here, we plot the downstream question-answering performance for the SMNetDecoder baseline (our best performing baseline from Sec. 5 in the main paper). We observe a 41\% drop in IoU on adding noise and evaluating using the original model. However, we see a 19\% increase on the same metric when we retrain SMNet on noisy data.

\xhdr{3. Question answering (QA) performance of our method (Fig. \ref{fig:noise-retraining} (c))}: Finally, we plot the downstream question-answering performance for our proposed method. Initially, on integrating noise and evaluating using the original model, we observe a 39\% drop in IoU performance. However, upon retraining SMNet and LingUNet, we see a 10\% increase on the same metric.

Across all the three experiments, we observe that (a) when models trained using privileged, oracle pose information are evaluated with noise in pose, their performance (understandably) drops and (b) we are able to recover the drop by re-training so that the model learns to adapt to its noisy inputs. We also qualitatively demonstrate the improvement in semantic map predictions obtained through a re-training of our models to adapt to noisy pose in Fig. \ref{fig:qual-retraining} (note the sharper boundaries and reduced label splatter).